\documentclass{IEEEtran}
\usepackage{amsmath,amsfonts}
\usepackage{algorithmic}
\usepackage{algorithm}
\usepackage{booktabs}
\usepackage{array}
\usepackage[caption=false,font=normalsize,labelfont=sf,textfont=sf]{subfig}
\usepackage{textcomp}
\usepackage{stfloats}
\usepackage{url}
\usepackage{verbatim}
\usepackage{graphicx}
\usepackage{enumerate}
\usepackage{caption}
\usepackage{cite}
\usepackage{epstopdf}         
\usepackage{flushend}
\usepackage{hyperref}
\usepackage{doi}
\usepackage{array}
\usepackage{booktabs}
\usepackage{threeparttable}
\usepackage{listings}
\usepackage{color}
\usepackage{float}
\usepackage{multirow} 
\usepackage{diagbox}
\usepackage{enumitem}
\definecolor{dkgreen}{rgb}{0,0.6,0}
\definecolor{gray}{rgb}{0.5,0.5,0.5}
\definecolor{mauve}{rgb}{0.58,0,0.82}
\usepackage{tabularx}
\usepackage{lineno}

\begin{document}
\title{AGSENet: A Robust Road Ponding Detection Method for Proactive Traffic Safety}

\author{Ronghui Zhang, Shangyu Yang, Dakang Lyu, Zihan Wang, Junzhou Chen, Yilong Ren, Bolin Gao, and Zhihan Lv

\thanks{Our manuscript was first submitted to IEEE Transactions on Intelligent Transportation Systems in June 30, 2023; (Corresponding author: Junzhou Chen.)}

\thanks{Ronghui Zhang, Shangyu Yang, Dakang Lyu, Zihan Wang and Junzhou Chen are with the Guangdong Provincial Key Laboratory of Intelligent Transport System, School of Intelligent Systems Engineering, Sun Yat-sen University, Guangzhou 510275,
China; (email: zhangrh25@mail.sysu.edu.cn; yangshy69@mail2.sysu.edu.cn;lvdk@mail2.sysu.edu.cn;wangzh579@mail2.s-ysu.edu.cn;chenjunzhou@mail.sysu.edu.cn).}
\thanks{Yilong Ren is with the School of Transportation Science and Engineering, Beihang University, Beijing 102206, China (e-mail: yilongren@buaa.edu.cn).}
\thanks{Bolin Gao is with the School of Vehicle and Mobility, Tsinghua University, Beijing 100084, China (e-mail: gaobolin@tsinghua.edu.cn).}
\thanks{Zhihan Lv is with the Department of Game Design, Faculty of Arts, Uppsala University, Uppsala 75236, Sweden (e-mail: lvzhihan@gmail.com).}

}

\markboth{IEEE Transactions on Intelligent Transportation Systems}%
{}

\maketitle

\begin{abstract}

Road ponding, a prevalent traffic hazard, poses a serious threat to road safety by causing vehicles to lose control and leading to accidents ranging from minor fender benders to severe collisions. Existing technologies struggle to accurately identify road ponding due to complex road textures and variable ponding coloration influenced by reflection characteristics. To address this challenge, we propose a novel approach called Self-Attention-based Global Saliency-Enhanced Network (AGSENet) for proactive road ponding detection and traffic safety improvement. AGSENet incorporates saliency detection techniques through the Channel Saliency Information Focus (CSIF) and Spatial Saliency Information Enhancement (SSIE) modules. The CSIF module, integrated into the encoder, employs self-attention to highlight similar features by fusing spatial and channel information. The SSIE module, embedded in the decoder, refines edge features and reduces noise by leveraging correlations across different feature levels. To ensure accurate and reliable evaluation, we corrected significant mislabeling and missing annotations in the Puddle-1000 dataset. Additionally, we constructed the Foggy-Puddle and Night-Puddle datasets for road ponding detection in low-light and foggy conditions, respectively. Experimental results demonstrate that AGSENet outperforms existing methods, achieving IoU improvements of 2.03\%, 0.62\%, and 1.06\% on the Puddle-1000, Foggy-Puddle, and Night-Puddle datasets, respectively, setting a new state-of-the-art in this field. Finally, we verified the algorithm's reliability on edge computing devices. This work provides a valuable reference for proactive warning research in road traffic safety.

\end{abstract}


\section{Introduction}
\begin{figure}
    \centering
    \includegraphics[scale=0.278]{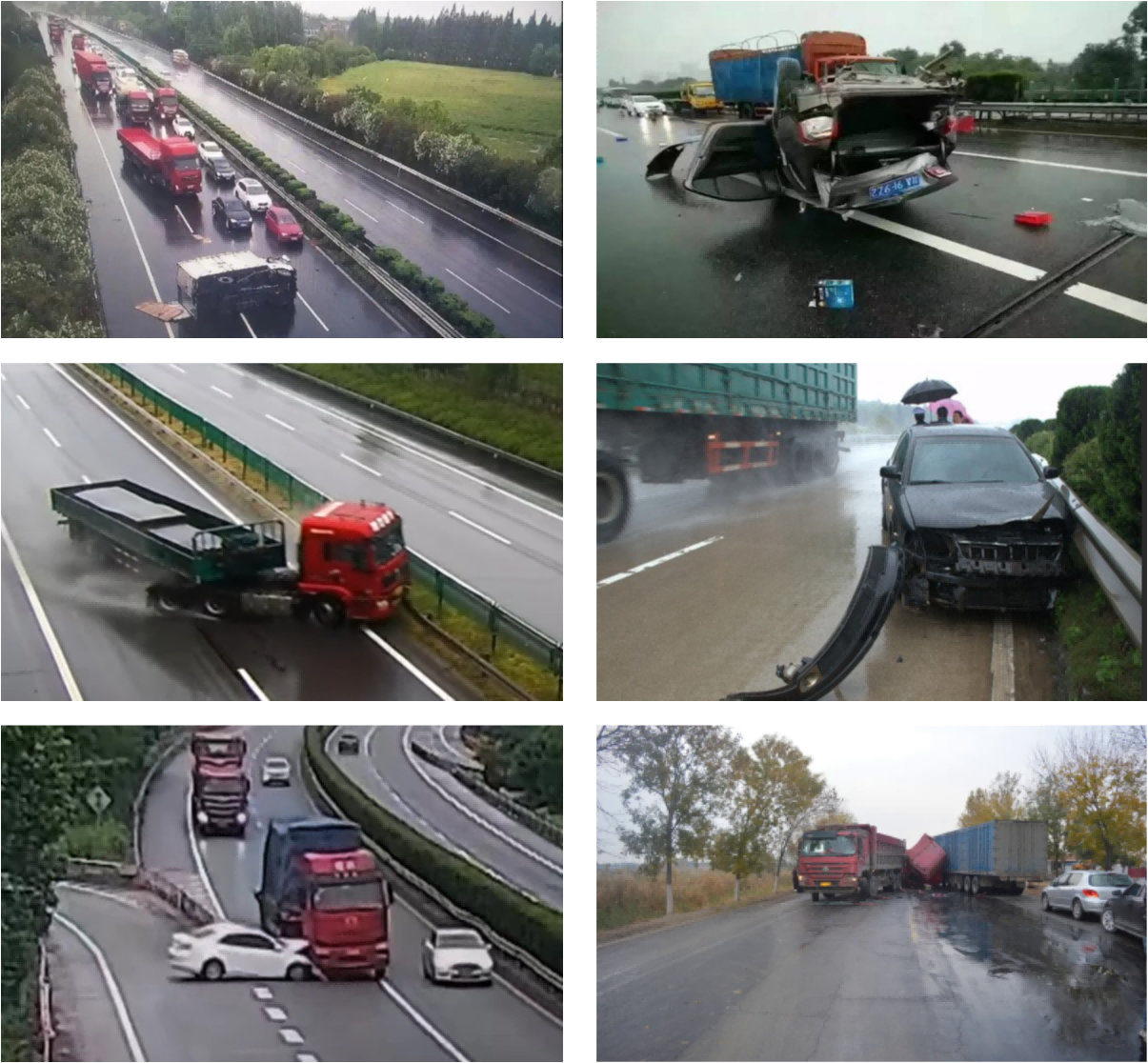}
    \caption{Typical traffic accidents caused by road ponding, from top to bottom, they are rollovers, skidding, and brake failure \cite{91,92,93,94,95,96}.}
    \label{fig:Typical}
\end{figure}

\IEEEPARstart{R}{oads} are intricate, large-scale systems where local anomalies can drastically disrupt the system's normal functioning \cite{1}. Road ponding, in particular, poses significant hazards. Specifically, as the vehicle wades through ponding traction between the tires and the road is reduced, causing the vehicle to lose control and skid. As a result, driving becomes more challenging and the risk of accidents increases (Fig. \ref{fig:Typical} displays typical accidents attributable to road ponding). Investigations suggest that inadequate warning about ponding areas is a principal factor in many vehicle collision accidents \cite{89}. Due to limited visibility, drivers often overlook or misjudge road ponding areas, particularly when driving at high speeds or tailgating. Consequently, the need for robust road ponding detection method to enhance traffic safety is pressing.

Current research often relies on special sensors, like vehicle-mounted laser scanners \cite{34}, acoustic sensors \cite{35}, ultrasonic sensors \cite{36,37}, optical spatial sensors \cite{41}, multi-spectral scanners \cite{24}, and near-field radars \cite{27}, to detect road ponding based on differing physical characteristics under varying road conditions. These methods can provide real-time information but require additional hardware and costs, and they fall short in subtle road ponding detection. With the development and application of image processing technology in road state recognition \cite{26,39,42,43}, image feature-based methods \cite{11,12,13} have offered new possibilities and research directions for road ponding detection. However, these methods are sensitive to lighting conditions and noise in real-world scenarios, demanding better robustness.

In the context of the rapid evolution and popularization of Deep Learning (DL), DL-based techniques have become the preferred choice for road state perception \cite{1,22,23}. Seminal work by FCN \cite{81} in the field of segmentation and subsequent studies \cite{31,32,33} utilizing the ``encoder-decoder" convolutional neural network design have made significant strides in road ponding detection \cite{58,57,61}. These models generate pixel-level probability maps, accurately identifying ponding areas and achieving impressive performance. Although semantic segmentation methods have accomplished competitive performance in road ponding detection, the task is still challenging due to road ponding's morphological, color-changing, visually ambiguous, and semi-transparent features.

The salient characteristic of road ponding is the reflection, causing complexities in the background and posing significant challenges for road ponding detection. Effectively leveraging this characteristic, instead of merely ignoring it, can significantly improve road ponding detection performance. Certain studies \cite{58,57} have leveraged reflection attention units for road ponding detection, enhancing performance at the cost of substantial computational resources.


Following the above-mentioned analysis, the necessity of employing richer saliency information for efficient road ponding detection is evident. With this motivation, we put forth the Self-Attention-based Global Saliency Enhancement Network (AGSENet), a novel road ponding detection approach anchored in salient object detection principles. In pursuit of more robust saliency feature learning in road ponding, we introduce the Channel Saliency Information Focus (CSIF) and Spatial Saliency Information Exploration (SSIE) modules. The CSIF module, seamlessly integrated into the encoder, employs a self-attention mechanism to explore the interrelationship among channels. Furthermore, it adeptly fuses spatial and channel information to accentuate features that exhibit significant similarity. Contrastingly, the SSIE module, skilfully incorporated into the decoder, utilizes spatial self-attention to delve into the correlations across different levels of feature, refining edge features and reducing noise, thus bridging the feature disparity across levels. Experimental validation underscores the superior performance of AGSENet in road ponding detection, significantly advancing the state-of-the-art. Considering the limited visibility caused by low light and foggy conditions, detecting road ponding becomes challenging and increases the risk of traffic accidents. To address this issue, we constructed the low-light-puddle and foggy-puddle datasets and conducted experiments to validate the generalizability and robustness of our method across different scenarios. The results demonstrate that our approach achieved state-of-the-art (SOTA) performance on these datasets. The key contributions of this paper are as follows:

\begin{enumerate}
\item \textbf{Salient Reflection Features:} To improve detection accuracy despite texture and color variations in ponding water, we introduced salient reflection features and developed AGSENet. The network's CSIF module uses self-attention to capture long-range similarity information, enhancing both spatial and channel saliency, which significantly boosts detection accuracy.

\item \textbf{Spatial Self-Attention Fusion:} We designed the SSIE module to address ambiguous features and detail loss during feature fusion. By leveraging high-level semantic information to suppress misleading features and refining edge details, this module bridges feature differences, enhances salient information, and reduces misdetections and omissions.

\item \textbf{Normalized Pixel Accuracy (NPA) Metric:} We introduced the NPA metric to address the limitations of the pixel-level accuracy (PA) metric in datasets with imbalanced classes. NPA uses rates of true and false positives and negatives, providing a more effective evaluation for imbalanced datasets.

\item \textbf{Dataset Relabeling and Expansion:} We relabeled the Puddle-1000 dataset using EISeg and created Night-Puddle and Foggy-Puddle datasets for low-light and foggy conditions. AGSENet showed IoU improvements of 2.03\%, 1.06\%, and 0.62\% and NPA increases of 0.46\%, 1.28\%, and 0.68\% on these datasets, demonstrating its robustness and generalization.
\end{enumerate}

The remaining sections of the paper are structured as follows: Section \uppercase\expandafter{\romannumeral2} provides a review of the related work. Section \uppercase\expandafter{\romannumeral3} outlines our proposed network architecture and loss function. Section \uppercase\expandafter{\romannumeral4}  presents the implementation details, evaluation metrics, experimental results and computational cost. Finally, Section \uppercase\expandafter{\romannumeral5} concludes the paper.

\begin{figure*}
     \centering
     \includegraphics[scale=0.6]{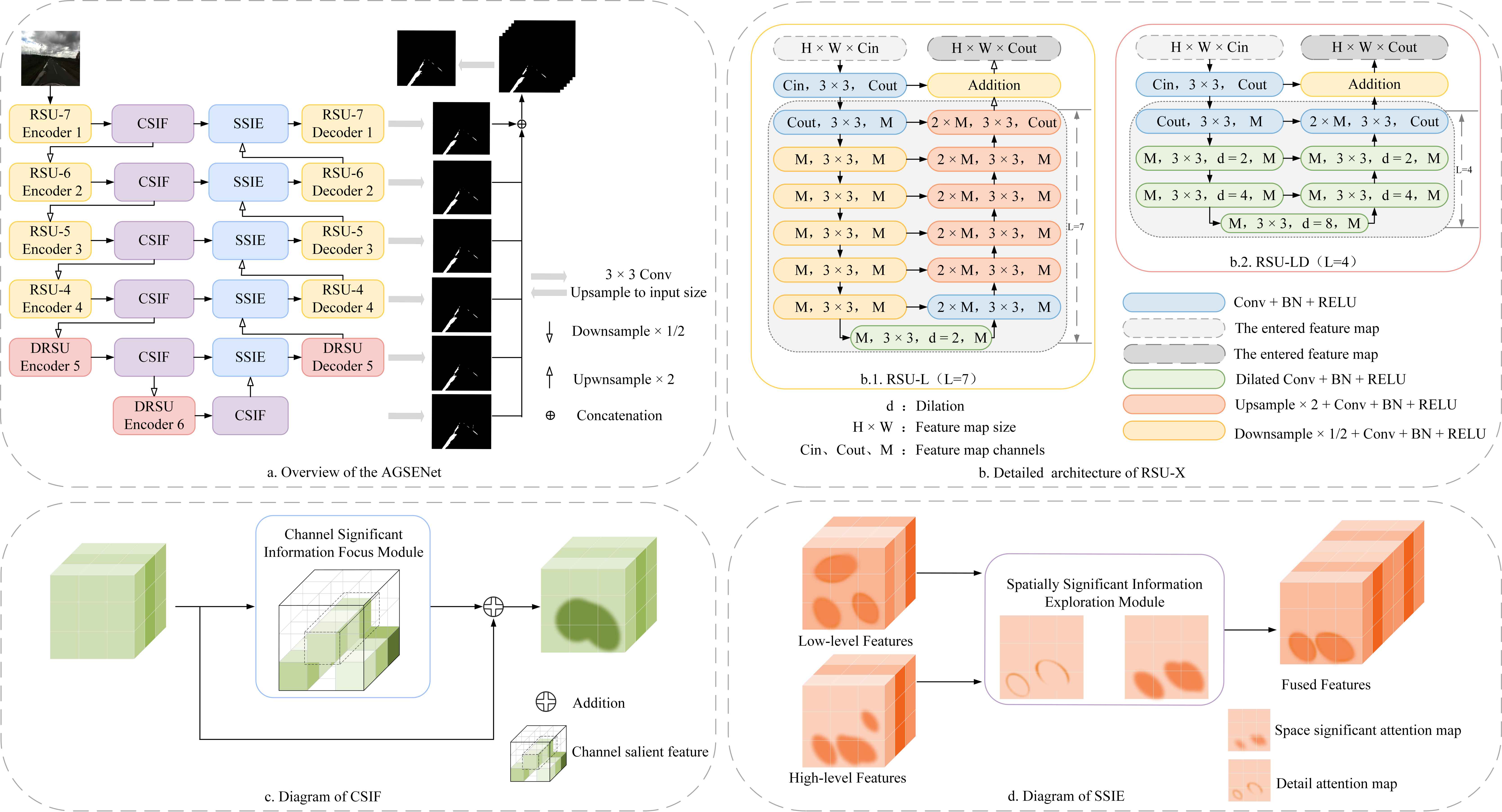}
     \caption{The framework of the proposed Self-Attention-based Global Saliency Enhancement Network (AGSENet) for Ponding-Water detection. The architecture incorporates an encoder, a decoder, a Channel Saliency Information Focus (CSIF) module (c), and a Spatial Saliency Information Exploration (SSIE) module (d). The encoder and decoder are built by stacking RSU-X (b), forming a U-shaped structure. The system is predicated on an encoder-decoder architecture to refine the segmentation of ponding water, with the CSIF module deployed to focus on salient information along the feature channel dimension. Concurrently, the SSIE module is harnessed to explore the spatial dimension's significant information across both high-level and low-level features.}
     \label{fig:AGSENet}
 \end{figure*}

\begin{figure*}
     \centering
     \includegraphics[scale=0.17]{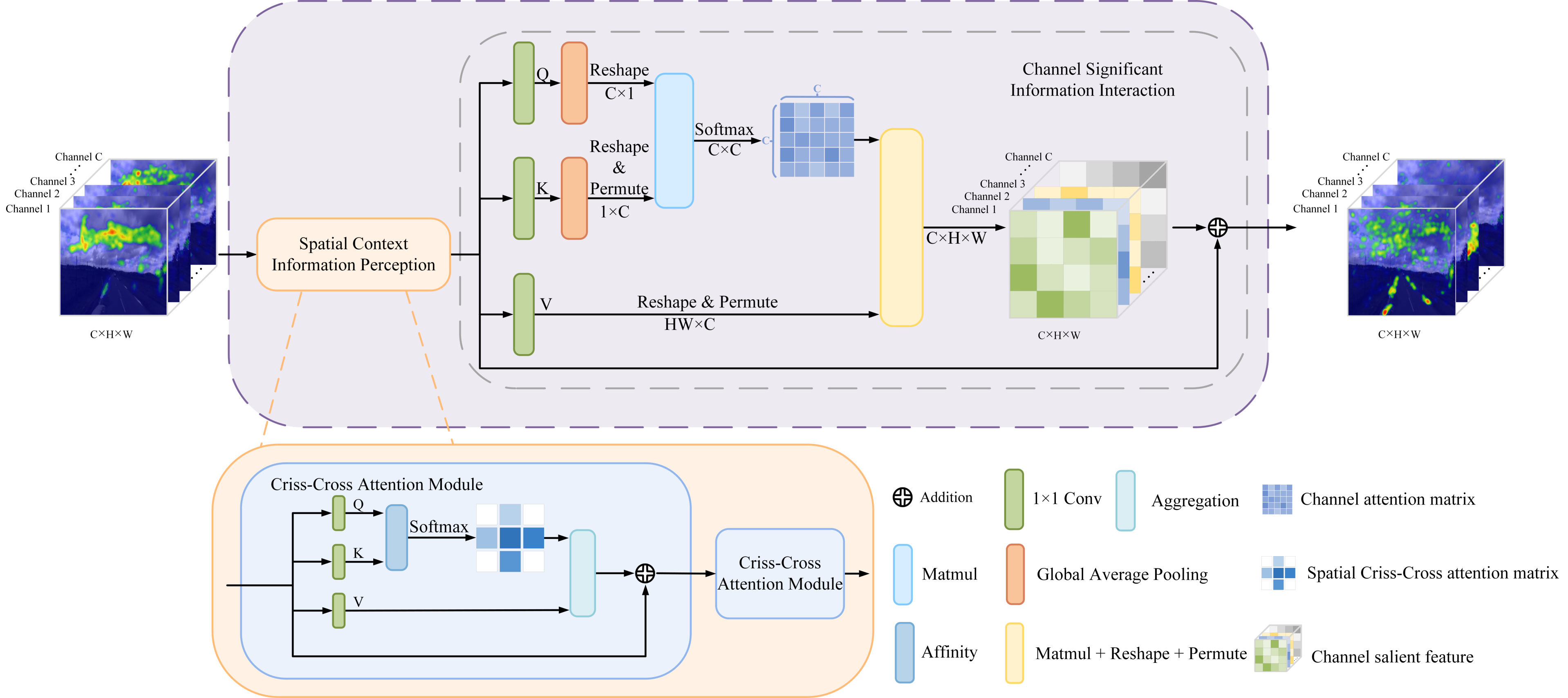}
     \caption{The structural diagram of the Channel Saliency Information Focus (CSIF) module includes Spatial Context Information Perception and Channel Saliency Information Interaction. The symbols Q, K, and V respectively represent Query, Key, and Value, while H, W, and C denote the shape of the features.}
     \label{fig:csif}
 \end{figure*}

\section{Related Work}
In this section, we provide a comprehensive overview of methodologies for road ponding detection, categorized into non-visual sensor-based approaches, traditional machine learning techniques, and deep learning methods. We also discuss the relevance of salient object detection techniques, which, although traditionally applied to different problems, offer valuable insights and solutions for the challenges in road ponding detection.

\subsection{Non-visual Sensor-based Methods}
In the field of road condition monitoring, various non-visual sensor-based methodologies have been explored for detecting standing water and other surface conditions. Study \cite{34} utilized a vehicle-mounted laser scanner to analyze the reflectivity of scattered echoes, enabling the identification of road conditions such as standing water. Research\cite{36} employed ultrasonic sensors to detect ponding water by analyzing changes in reflection patterns due to surface roughness variations. Another study \cite{37} focused on the disparities in echo impedance at different ultrasonic frequencies to identify areas with standing water. Acoustic sensors were leveraged in study \cite{35} to detect standing water in urban and highway settings by extracting acoustic features and performing sub-band frequency analysis. Additionally, research \cite{38} developed a complex impedance-based sensor capable of distinguishing between dry roads, ice, and standing water by examining the capacitance and conductivity properties of the road surface materials. While these non-visual sensor methodologies provide real-time road condition data, they often require additional hardware and incur higher costs. Furthermore, their effectiveness in detecting subtle instances of road ponding is limited, which constrains their practical application in real-world scenarios.

\subsection{Machine Learning-based Methods}
\subsubsection{Traditional Machine Learning-based Methods}
Traditional machine learning techniques for road surface state detection focus on manually extracted features such as color, texture, and brightness. Common algorithms include Support Vector Machines (SVM) \cite{52,48}, K-nearest neighbors (KNN) \cite{43}, and Bayesian classifiers \cite{42}.

For instance, PCA was employed in \cite{39} to transform texture features for nighttime wet/dry surface detection using Mahalanobis distance. A continuous frame reflection model rapidly identified surface states in \cite{42}. Integration of color, brightness, and texture features with KNN aided detection under varying lighting conditions \cite{43}. Brightness ratios across wavelengths were used in \cite{44} for pixel state classification employing KNN, SVM, and Partial Least Squares. Wavelet transforms facilitated surface state extraction in \cite{48}, classified by SVM.

Optimized SVM and image segmentation identified ponding areas in \cite{45}. Gradient RGB color features trained SVM models in \cite{46}, and brightness normalization assisted SVM and Bayesian classifiers in \cite{47}. Near-infrared images combined with PCA detected waterlogged areas in \cite{49}. Optical flow integrated with SVM enabled small-scale ponding detection in \cite{52}, while \cite{50} tracked ponding reflections considering polarized light.

These traditional methods, while robust and cost-effective, often struggle in complex environments and varying light conditions, necessitating enhanced anti-interference capabilities.

\subsubsection{Deep Learning-based Methods}
Deep learning has transformed road surface state detection, offering high accuracy and generalization by learning features directly from data. Methods typically involve object detection and image segmentation using Convolutional Neural Networks (CNNs) and other advanced architectures.

For example, \cite{56} proposed a CNN-based method integrating road friction coefficients to classify six surface types, including ponding. Reflection attention modules in Fully Convolutional Networks (FCN) enhanced ponding reflection detection in \cite{58}. Conditional Generative Adversarial Networks (cGAN) with reflection attention improved water hazard detection in \cite{57}. A CNN approach in \cite{59} provided 3D visualization of ponding detection results. ResNet50 identified waterlogged urban roads in \cite{60}, while SegNet segmented water splashes on highways in \cite{61}.

Preprocessing techniques like data splitting and enhancement supported a CNN-based detection method in \cite{62}. The RCNet model, based on the CNN framework, achieved accurate classification across various road conditions in \cite{63}.

The success of the Transformer architecture\cite{need} in natural language processing (NLP) has inspired the exploration of its application in visual tasks, demonstrating potential performance improvements. Vision Transformers (ViT) \cite{vit} performs well in some computer vision tasks. By dividing images into small blocks and processing the relationship between these blocks, ViT is able to capture global contextual information, which is particularly important for detecting water with complex reflective properties. However, the application of Transformers in road water detection is still rare and needs further research.

While traditional machine learning and deep learning methods have shown effectiveness in detecting road ponding, they often face challenges in accurately identifying ponding areas due to their reliance on general image features. Salient Object Detection (SOD) techniques, which aim to highlight the most visually prominent objects in an image, provide an alternative approach. By focusing on saliency features such as color, texture, and contours, SOD can enhance the detection of road ponding, which is often characterized by complex reflective properties.

\subsection{Salient Object Detection}
Salient object detection (SOD) aims to segment prominent objects at the pixel level using features such as color, texture, and contours. This task is crucial for identifying primary subjects in images, with applications across various domains \cite{68,69,70,71,72,73,74}. In this study, road ponding, characterized by complex visual features due to reflective phenomena, is treated as a salient object, making its detection a specific SOD task.

Traditional SOD methods leverage low-level visual features and saliency priors, including center bias \cite{75}, scene contrast \cite{76}, foreground-background \cite{77}, and edge modulation \cite{28}. Notable examples are the Dense Sparse Labeling (DSL) framework \cite{14} and superpixel-level graphs incorporating color and motion histograms \cite{87}. Despite their utility, these methods rely on manually defined features, limiting their robustness and generalization in complex scenarios such as road ponding.

The introduction of Convolutional Neural Networks (CNNs) has revolutionized SOD by enabling the extraction of multi-scale and multi-level features, enhancing the detection accuracy of salient regions like road ponding. Techniques such as iterative optimization of multi-scale features \cite{18}, self-interacting modules to reduce cross-scale noise \cite{29}, and the cascaded feedback decoder in F3Net \cite{20} have significantly advanced the field. The Residual U (RSU) architecture in U2Net \cite{79} improves contextual information capture, while other methods \cite{17,19} address interference and integrate multi-task learning paradigms.

Recent innovations include cross-modal fusion models using Swin Transformers for hierarchical feature extraction and attention mechanisms \cite{6}, and dual-attention residual networks to refine object parts and boundaries \cite{5}. Additionally, the grafting strategy in \cite{30} combines CNN and Transformer features for high-resolution segmentation.

Our proposed approach introduces a saliency feature enhancement module based on self-attention mechanisms, specifically tailored for road ponding detection. This method emphasizes salient features while preserving edge details, ensuring robust and efficient detection in diverse driving scenarios.

\section{Method}
\subsection{Motivation} 
Detecting road ponding typically involves analyzing image attributes, including color features (such as chroma, brightness, and saturation) and composite features (like color and texture). Existing detection strategies, such as laser radar-based methods \cite{34}, machine learning-based methods \cite{45}, and image-based methods \cite{22,23}, often struggle with robust detection of road ponding areas, particularly subtle ponding. High computational complexity and failure to satisfy the safety warnings' requirements in road supervision systems further exacerbate these issues. The authors have comprehensively researched standing water features to enhance road ponding detection. We identified the potential benefits of appropriately utilizing the reflective properties of road ponding instead of disregarding or removing them \cite{58}. Therefore, this study aims at learning salient features rooted in the reflection phenomenon under complex driving scenarios to attain robust road ponding detection. Accordingly, we propose a saliency feature enhancement method exploiting the self-attention mechanism. This method applies the Channel Saliency Information Focus (CSIF) module to capture significant similarity features instigated by the reflection phenomenon. Subsequently, it fuses different-level features through the Space Saliency Information Exploration (SSIE) module. Integrating these modules, we incorporate the saliency feature enhancement method into the U2Net \cite{79} framework, culminating in the Self-Attention-based Global Saliency Enhancement Network (AGSENet). The result is a more robust road ponding detection in complex driving scenarios, improving the safety warnings in road supervision systems.
\subsection{Overall} 
As depicted in Fig. \ref{fig:AGSENet}, our architecture capitalizes on the U2Net \cite{79} framework to discern the unique visual characteristics of road ponding. However, we augment the road ponding saliency features by integrating the Channel Salient Information Focus (CSIF) module and the Spatial Salient Information Exploration (SSIE) module within the U2Net. An RGB image is processed in several steps within the network. First, the image is input into the network's encoder to extract multi-scale features. We embed the CSIF module after each RSU-X block of the encoder to learn more comprehensive representations of road ponding saliency features, thereby enhancing feature robustness. Next, we employ the SSIE module during feature fusion to achieve two key goals: ambiguous feature rejection and edge feature refinement. This ensures that the final fused features offer high discriminability and accurately encapsulate road ponding's salient features. The fused features are then passed to the network's decoder, which uses progressive upsampling and convolution operations to restore the high-resolution feature map and refine the segmentation of road ponding. Finally, to enrich multi-scale feature information, we perform convolution and upsampling operations on Encoder 6's output and Decoder 1-5. We generate saliency maps of the exact dimensions as the input image and combine them using a cascade operation. A 1×1 convolution is applied lastly to generate the final prediction map for road ponding. The subsequent sections delve into a detailed explanation of the Encoder-Decoder structure, CSIF module, and SSIE module of our network.

\subsection{Encoder-Decoder}
The AGSENet architecture leverages a U-shaped encoder-decoder framework comprised of a stack of several RSU-X blocks. This design facilitates effective multi-scale feature extraction and learning, enhancing segmentation performance. In this section, we will separately discuss the RSU-X block and the encoder-decoder structure of the AGSENet.

The specific structure of the RSU-X block, shown in Fig. \ref{fig:AGSENet} (b), consists of two versions: RSU-L block (Fig. \ref{fig:AGSENet} (b.1)) and RSU-LD block (Fig. \ref{fig:AGSENet} (b.2)). The RSU-L block is a U-Net-like encoder-decoder structure containing L layers, where the size is determined by the input image resolution. Each RSU-L block contains three components: (1) The input convolution layer, responsible for local feature extraction, which transforms the input feature map $ X \in \mathbb{R}^{C_{in} \times H \times W} $ into a feature $F_1(x)\in \mathbb{R}^{C_{out} \times H \times W} $; (2) The encoder-decoder structure, responsible for extracting and fusing multi-scale features. This structure takes the intermediate feature map $F_1(x)$ as input, extracts multi-scale features via progressive pooling and convolution operations, and subsequently restores them to high-resolution feature maps through progressive upsampling, concatenation, and convolution operations, ultimately producing fused multi-scale features $F_2(x)\in \mathbb{R}^{C_{out} \times H \times W} $; (3) The residual connection, which fuses local features and multi-scale features $F_3(x)\in \mathbb{R}^{2C_{out} \times H \times W} $ through a concatenate operation. The formula for the RSU-L process is expressed as:

\begin{align}
&F_{1}(X)=R\left(N\left(C_{3 \times 3}(X)\right)\right), \\
&F_{2}(X)=U\left(F_{1}(X)\right), \\
&F_{3}(X)=\operatorname{Cat}\left((F_{1}(X), F_{2}(X)), dim=1\right),
\end{align}

Where $C_{3 \times 3}$ represents the convolution layer with a kernel size of 3; $ U $ represents the bilinear upsampling; $ N $ represents the batch normalization (BN); $ R $ represents the ReLU activation function; and $ Cat $ represents the concatnate operation.

To overcome the loss of contextual information caused by downsampling at lower resolution input feature maps, U2Net introduces the RSU-LD block as an extension of the RSU-L block. In the RSU-LD block, L represents the number of encoder layers. Unlike the RSU-L block, which employs normal convolutions, pooling, and upsampling operations, the RSU-LD block utilizes dilated convolutions. By leveraging dilated convolutions, the RSU-LD block enhances the network's ability to capture detailed information and improve the segmentation performance.

\begin{table}[h]
\centering
\caption{The detailed configuration of the encoder-decoder of AGSENet,“$C_{in}$”, “$M$”and “$C_{out}$” indicate input channels, middle channels and output channels.}
\label{set}
\tabcolsep=0.3cm
\renewcommand\arraystretch{1.5}
\begin{tabular}{c|c|c|c|c|c|c}
\hline
\multirow{2}{*}{RSU-X} &\multicolumn{3}{c|}{Encoder 1-6} &\multicolumn{3}{c}{Decoder 1-5} \\
\cline{2-7}
& $C_{in}$ & $M$ & $C_{out}$ & $C_{in}$ & $M$ & $C_{out}$ \\
\cline{1-7}
RSU-7	&3	&16	&64	&128	&16	&64 \\ 

RSU-6	&64	&16	&64	&128	&16	&64 \\

RSU-5	&64	&16	&64	&128	&16	&64 \\

RSU-4	&64	&16	&64	&128	&16	&64 \\

RSU-4D	&64	&32	&128	&256	&32	&64 \\

RSU-4D	&128	&32	&128	&\diagbox{}{}	&\diagbox{}{}	&\diagbox{}{} \\
\hline
\end{tabular}
\label{table_MAP}
\end{table}
The detailed composition of AGSENet's encoder-decoder architecture is presented in Table \ref{set}. This architecture is composed of three main elements: (1) A six-stage encoder, (2) A five-stage decoder, and (3) The saliency map fusion module, which bridges Decoder 1-5 and Encoder 6. The functionalities of these components are as follows:
(i) Encoder stages (Encoder 1-4): These stages process high resolution feature maps. RSU-7, RSU-6, RSU-5, and RSU-4 are utilized in Encoder 1, Encoder 2, Encoder 3, and Encoder 4, respectively, to capture large-scale information. For Encoder 5 and Encoder 6, which process relatively lower resolution feature maps, RSU-4D is employed.
(ii) Decoder stages (Decoder 1-5): The structure of the decoder stages is similar to that of the encoder stages, with RSU-4D being employed in Decoder 5.
(iii) Saliency map fusion module: In this module, the outputs of Encoder 6 and Decoder 1-5 undergo convolution and upsampling operations to obtain saliency maps that match the input image's resolution. These maps are then concatenated, and a final prediction map is generated through a 1×1 convolution operation.
\begin{figure}
    \centering
    \includegraphics[scale=0.14]{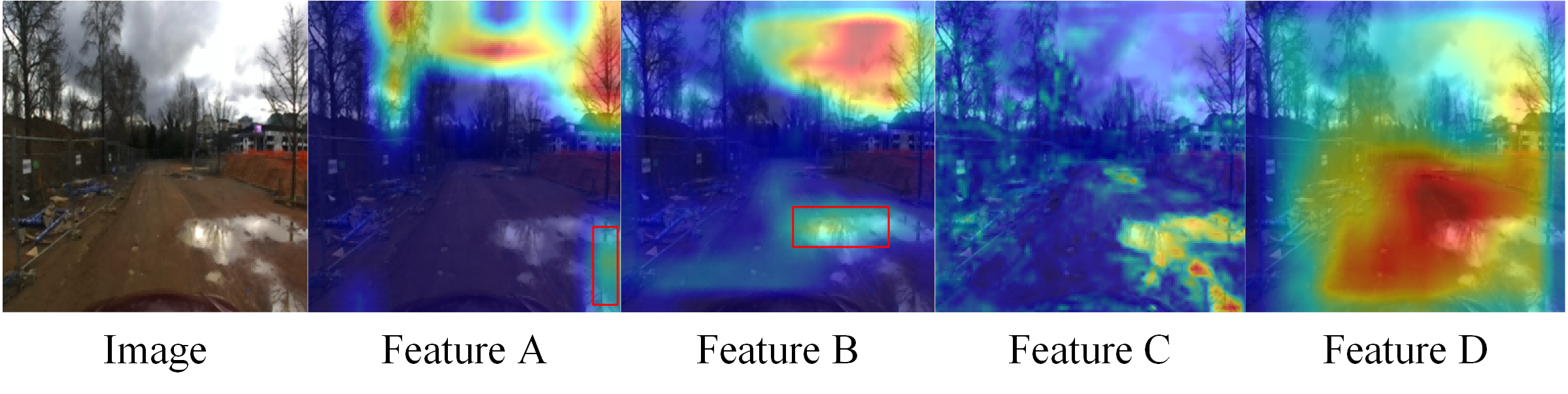}
    \caption{The responses of different channels to specific object categories are shown from left to right as trees, sky, water, and ground, respectively. The red boxes represents the reflection of objects in the water in a real-world scene.}
    \label{fig:feature}
\end{figure}
\begin{figure*}
    \centering
    \includegraphics[scale=0.15]{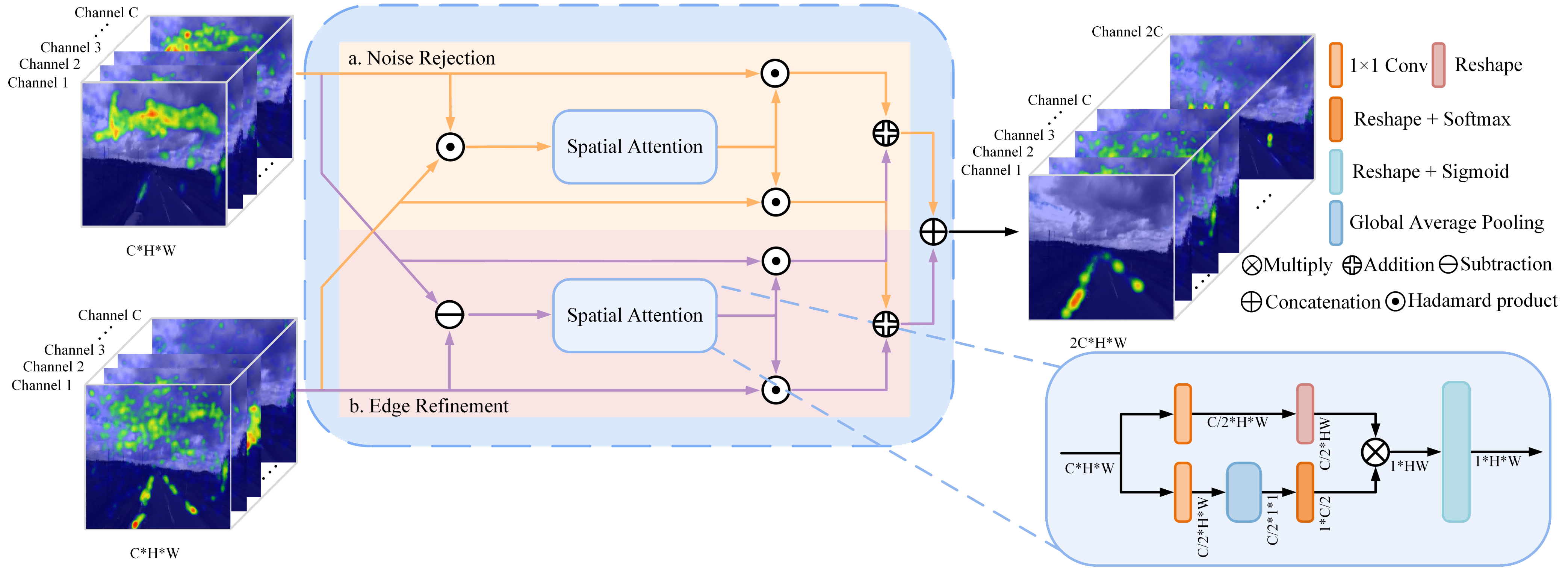}
    \caption{The structural diagram of the Spatial Saliency Information Exploration (SSIE) module includes ambiguous feature noise removal and edge position refinement. H, W, and C denote the shape of the features.}
    \label{fig:HLE}
\end{figure*}
\begin{figure*}
    \centering
    \includegraphics[scale=0.47]{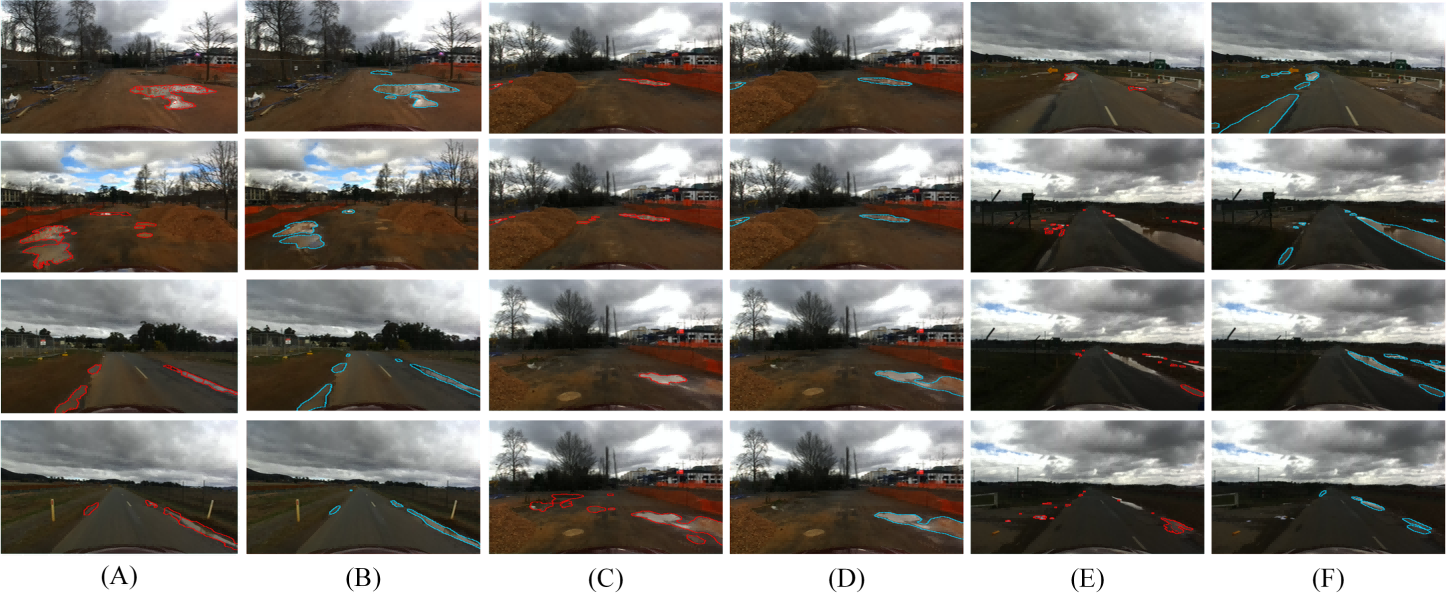}
    \caption{ For clarity, we only emphasize the mask edges. Among them, A arises from discontinuity and cluttered edges induced by reflection; C results from inconsistent labeling of context road ponding; E results from missed or erroneous labeling; B, D, and F illustrate the corrections for the above three scenarios.}
    \label{fig:data}
\end{figure*}

\subsection{Channel Saliency Information Focus Module}
In neural networks, each channel map indicates a specific response to a particular characteristic or category within the input image, as demonstrated in Fig. \ref{fig:feature}. The reflection caused by road ponding can result in high responses across multiple channels due to the presence of information from other categories in the water pixels. Instead of increasing redundancy, our goal is to utilize channel-level information to capture the salient features of water and enhance the representation of specific semantic features. To achieve this, we propose the Channel Saliency Information Focus (CSIF) module, which explicitly models the significance between channels, as depicted in Fig. \ref{fig:csif}.

Unlike existing channel attention mechanisms \cite{78,80}, our approach incorporates an adaptive aggregation strategy to effectively combine long-range significant information, thus enhancing feature representation for road ponding without introducing unnecessary redundancy. The CSIF module comprises two stages: Spatial Context Information Perception (SCIP) and Channel Significant Information Interaction (CSII).

Initially, we utilize the output of adjacent stage encoders to input the CSIF module and capture richer contextual information through the Spatial Context Information Perception (SCIP). This process employs Criss-Cross Attention \cite{82}, albeit with a difference: we collapse the Q and K dimensions entirely along the channel axis, reducing the number of channels to 1. This approach simplifies the model, decreases the number of parameters, and its detailed structure can be found in \cite{82}.

Subsequently, we aggregate long-range significant information along the channel axis through Channel Significant Information Interaction (CSII). This process is delineated in the following steps:

(1) We apply three convolutional layers with 1×1 filters to the input feature map $X \in \mathbb{R}^{C \times H \times W}$, generating three feature maps $Q$, $K$, and $V$, where {$Q$, $K$, $V$} $\in \mathbb{R}^{C \times H \times W}$. Subsequently, we perform a complete collapse of the spatial dimensions of $Q$ and $K$ via global average pooling, encapsulating the spatial information into the channel descriptor, yielding two feature maps $Q'$ and $K'$, where {$Q'$, $K'$} $\in \mathbb{R}^{C \times 1 \times 1}$. Due to the employment of Spatial Context Information Perception, the representational capability of spatial dimension pixels is strengthened, thus mitigating the information loss attributed to the spatial dimensions collapse:

\begin{align}
&Q'=Gap\left(C_{1 \times 1}\left(X\right)\right), \\
&K'=Gap\left(C_{1 \times 1}\left(X\right)\right), \\
&V=C_{1 \times 1}\left(X\right),
\end{align}

Where $Gap$ is the global average pooling operation; $C_{1 \times 1}$ is the convolution layer with a kernel size of 1.

(2) We reshape the globally averaged pooled $Q$ and $K$ into $\mathbb{R}^{C \times 1}$ and $\mathbb{R}^{1 \times C}$, respectively, execute matrix multiplication, and subsequently apply a softmax layer to derive the channel similarity attention map $A \in \mathbb{R}^{C \times C}$:

\begin{align}
&a_{ij} = \frac{exp\left(Q'_i \cdot K'_j\right)}{\sum_{i=1,j=1}^C exp\left(Q'_i \cdot K'_j\right)},
\end{align}

where $a_{ij}$ quantifies the influence of the $i^{th}$ channel on the $j^{th}$ channel. The correlation between two positions is directly proportional to the similarity of their feature representations.

(3) We reshape $V$ obtained in (1) into $\mathbb{R}^{C \times N \times 1}$, where $N = H × W$ is the number of pixels, fuse spatial and channel information by performing matrix multiplication between $V$ and $X$, obtaining $ P \in \mathbb{R}^{C \times N} $, and then reshape the result to obtain the aggregated salient feature map of road ponding. To address the issue of vanishing gradients, we multiply the obtained result by a scale parameter and then perform an element-wise summation with x to obtain the final output of the CSIF module $ F \in \mathbb{R}^{C \times H \times W} $.

\begin{align}
&p_{ij}=\sum_{k=1}^C \left(a_{ik}\cdot v_{kj}\right), \\
&F=\alpha\left(Reshape\left(P\right)\right)+X,
\end{align}

where $p_{ij}$ represents the element in the $i^{th}$ row and $j^{th}$ column of matrix $P$; $a_{ik}$ signifies the element in the $i^{th}$ row and $k^{th}$ column of matrix $A$; $v_{kj}$ corresponds to the element in the $k^{th}$ row and $j^{th}$ column of matrix $V$.

Diverging from the approach presented in \cite{90}, we incorporate spatial information into channel descriptors and facilitate interactions with channel information, thereby enabling the synchronization of global contextual information. Moreover, our methodology markedly diminishes the number of parameters, ensuring efficiency while maintaining accuracy.

\subsection{Spatial Saliency Information Exploration Module}
Neural networks, with their varying hierarchies and filter sizes, are adept at capturing both high-level and low-level features. Low-level layers in neural networks have small receptive fields, which make them proficient in capturing fine-grained details such as edges, lines, and corners. However, the reflective properties of road ponding often cause low-level layers to exhibit high responses for many background (non-water) categories, leading to substantial noise in the features. On the other hand, deeper layers, with larger receptive fields, are capable of capturing and integrating information from broader regions of the input space, thus encoding high-level semantic information and understanding contextual relationships. However, this broad perspective often comes at the expense of detailed, localized information.

Therefore, simple fusion of features from different levels can introduce blurred features (background noise that bears high similarity to road ponding features) or result in the loss of critical details, hindering the network's optimization. To address this issue, we designed the Spatial Saliency Information Exploration (SSIE) module to bridge the feature gap between different tiers and to enhance significant spatial information associated with road ponding. As illustrated in Fig. \ref{fig:HLE}, the SSIE module accepts two inputs: the up-sampled feature map from the previous level decoder and the feature map from the symmetric encoder level. It processes these inputs and outputs fused features.

Specifically, we first feed the high-level feature $F_h \in \mathbb{R}^{C \times H \times W}$ and the low-level feature $F_l \in \mathbb{R}^{C \times H \times W}$ of the same shape into the Edge Refinement (ER) and Noise Rejection (NR) branches, respectively. We then execute pixel-wise subtraction and Hadamard operations to derive two feature maps, $F_n \in \mathbb{R}^{C \times H \times W}$ and $F_e \in \mathbb{R}^{C \times H \times W}$. Subsequently, we employ a spatial self-attention mechanism to derive spatial attention maps for $F_n$ and $F_e$, and then multiply them with $F_h$ and $F_l$, yielding $F_{h}^{n}$, $F_{h}^{e}$, $F_{l}^{n}$, and $F_{l}^{e}$. Finally, pixel-wise addition is performed at their respective levels, resulting in $F'_h$ and $F'_l$ feature maps, which are then concatenated to generate the fused feature $F_f \in \mathbb{R}^{2C \times H \times W}$:

\begin{equation}
    F_{h}^{k} =
\begin{cases} 
\mathit{SA}(F_{h}-F_{l}) \times F_{h}, & k = n \\
\mathit{SA}(F_{h} \times F_{l}) \times F_{h}, & k = e
\end{cases}
\label{head}
\end{equation}

\begin{equation}
    F_{l}^{k} =
\begin{cases} 
\mathit{SA}(F_{h}-F_{l}) \times F_{l}, & k = n \\
\mathit{SA}(F_{h} \times F_{l}) \times F_{l}, & k = e
\end{cases}
\label{head}
\end{equation}

\begin{align}
    &F'_h = F_{h}^{n} + F_{h}^{e}, \\
    &F'_l = F_{l}^{n} + F_{l}^{e}, \\
    &F_f = Cat(F'_h, F'_l, dim=1),
\end{align}

where $Cat$ denotes the concatenation operation and $SA$ represents the spatial self-attention.

The SSIE module differs from previous methods (e.g., \cite{88}) in that we utilize spatial self-attention mechanisms to obtain correlation attention maps, which guide the enhancement of spatial saliency features. Additionally, our method reduces the number of parameters, improving efficiency.
\subsection{Loss Function}
Segmentation of road ponding faces a formidable challenge due to significant class imbalance - the ponding water comprises a minor fraction of the image, while the majority constitutes non-water background. This imbalance can significantly impair the network's performance as the ponding water-related loss is often overwhelmed. Existing strategies to overcome this issue typically incorporate weight adjustments within the loss function, as exemplified by HED \cite{66} and Focal Loss \cite{64}.
This paper proposes a straightforward and efficient solution to the class imbalance problem in road ponding segmentation. We introduce a weighted hybrid loss method, which is formulated as follows:
\begin{align}
&L_{hybrid}=\gamma L_{BCE}+\delta L_{DICE}, 
\end{align}

Where $L_{BCE}$ and $L_{DICE}$ denote the binary cross-entropy ($BCE$) loss \cite{65} and $DICE$ loss \cite{67}, respectively. $\gamma$ and $\delta$ are the learnable scale parameters which are initialized as 1.

The BCE loss \cite{65}, widely utilized in binary classification and segmentation tasks, computes the loss independently for each pixel. This pixel-wise measurement is beneficial for achieving convergence across all pixels. Nevertheless, since BCE loss \cite{65} assigns identical weight to foreground and background pixels, it might dilute the loss for foreground pixels in scenarios where the background dominates, such as in road ponding segmentation. To remedy this issue and encourage the network to pay greater attention to the foreground region, we introduce the DICE loss \cite{67}. Additionally, we incorporate deep supervision into our loss function to facilitate network learning. Thus, our overall loss function is defined as follows:
\begin{align}
&L_{total}=\sum_{d=1}^D L_{hybrid}^{\left(d\right)}+L_{hybrid}^{\left(fuse\right)}, 
\end{align}

Where $L_{hybrid}^{\left(d\right)}$ ($D$=6) represents the loss of the saliency map generated by the Encoder 6 and Decoder 1-5. $L_{hybrid}^{\left(fuse\right)}$ represents the loss of the final fused output prediction map.
\begin{table*}
\centering
\caption{Quantitative comparison data with other methods are presented. † represents the experimental results reproduced by our team based on the original paper using the re-labeled Puddle-1000 dataset. Seg stands for segmentation, SOD represents salient object detection, and RPD signifies road ponding detection. The unit for $IoU$, $MIoU$, $PA$, $F_\beta$ is percentage ($\%$). The first and second best results are indicated in \textbf{\textcolor{blue}{blue}}, \textcolor{green}{green}, respectively.}
\label{ONR}
\tabcolsep=0.11cm
\renewcommand\arraystretch{1.5}
\begin{tabular}
{c|c|c|cccc|cccc|cccc}
\hline
\multirow{2}{*}{Methods} &\multirow{2}{*}{Pub.’Year} &\multirow{2}{*}{Task} &\multicolumn{4}{c|}{Puddle-1000(ONR)} &\multicolumn{4}{c|}{Puddle-1000(OFR)} &\multicolumn{4}{c}{Puddle-1000(MIX)}\\
\cline{4-15}
&&& $IoU$ & $MIoU$ & $NPA$ & $F_\beta$ & $IoU$ & $MIoU$ & $NPA$ & $F_\beta$ & $IoU$ & $MIoU$ & $NPA$ & $F_\beta$\\
\cline{1-15}
FCN8s-RAU \cite{37}†	&ECCV'2018	&RPD	&82.48	&91.16	&95.21	&90.39 &84.16 &91.98 &95.89 &91.39 &83.61 &91.71 &95.90 &91.07 \\

Deeplabv3+ \cite{93}†	&ECCV'2018	&Seg &81.46	&90.65	&94.98	&89.78 &83.02 &91.41 &95.44 &90.72 &82.67 &91.24 &95.69 &90.51 \\

CCNet \cite{ccnet}†	&IEEE TPAMI'2020 &Seg	&80.73	&89.27	&94.72	 &86.83 &83.52 &91.64 &95.81 &91.61 &83.10 &91.47 &95.75&90.76 \\ 

F3Net \cite{62}†	&AAAI'2020	&SOD	&77.54	&88.66	&94.17	&87.35 &83.91 &91.85 &95.81 &91.25 &80.28 &90.03 &94.97 &89.06 \\

U2Net \cite{71}†	&PR'2020	&SOD	&83.56&91.71	&95.72	&91.04 &83.26 &91.53 &95.68 &90.86 &82.97 &91.39 &95.84 &90.69 \\

Segformer-B2 \cite{94}†	&NeurIPS'2021	& Seg&81.92 & 90.89 & 95.24 & 90.86 & 84.24 & 92.03 & 95.87 &91.44 & 80.82 & 90.29 & 95.13 & 89.39\\ 

SWNet \cite{39}†	&IEEE TITS'2022	&RPD	&82.68	&91.26	&95.42	&90.52 &82.95 &91.37 &95.36 &90.68 &81.59 &90.69 &95.26 &89.86 \\

Seaformer-Base \cite{seaformer}†	&ICLR'2023	&Seg	& 82.57 & 91.21 & 95.30 & 90.48 & 84.40 &92.13 &95.91 &91.56 &82.88 &91.36 & 95.73 & 90.54\\ 

Homofusion \cite{homofusion}†	&ICCV'2023	&RPD	& 82.71 & 91.45 & 95.61 & 90.57 & \textcolor{green}{84.55} & \textcolor{green}{92.25} & \textcolor{green}{95.94} & \textcolor{green}{91.62} &83.65 & 91.74 & 95.94 &91.12 \\

GDNet \cite{gdnet}†	&IEEE TPAMI'2023	&SOD	& \textcolor{green}{83.62} & \textcolor{green}{91.75} & \textcolor{green}{95.78} & \textcolor{green}{91.14} & 84.10 & 91.87 & 95.84 & 91.34 & \textcolor{green}{83.81} & \textcolor{green}{91.74} & \textcolor{green}{96.22} & \textcolor{green}{91.16} \\ 

\textcolor{blue}{OURS}	&\textcolor{blue}{- - - -}	&\textcolor{blue}{RPD}	&\textbf{\textcolor{blue}{85.67↑}}	&\textbf{\textcolor{blue}{92.78↑}}	&\textbf{\textcolor{blue}{96.54↑}}	&\textbf{\textcolor{blue}{92.27↑}} &\textbf{\textcolor{blue}{85.68↑}} &\textbf{\textcolor{blue}{92.77↑}} &\textbf{\textcolor{blue}{96.76↑}} &\textbf{\textcolor{blue}{92.29↑}} &\textbf{\textcolor{blue}{85.84↑}} &\textbf{\textcolor{blue}{92.85↑}} &\textbf{\textcolor{blue}{96.68↑}} &\textbf{\textcolor{blue}{92.38↑}} \\
\hline
\end{tabular}
\label{ONR}
\end{table*}

\begin{table}[htbp]
\centering
\caption{Quantitative comparison data with other methods on the Foggy-Puddle are presented. Seg stands for segmentation, SOD represents salient object detection, and RPD signifies road ponding detection. The unit for $IoU$, $MIoU$, $PA$, $F_\beta$ is percentage ($\%$). The first and second best results are indicated in \textbf{\textcolor{blue}{blue}}, \textcolor{green}{green}, respectively.}
\label{foggy_methods}
\tabcolsep=0.03cm
\renewcommand\arraystretch{1.5}
\begin{tabular}{c|c|c|cccc}
\hline
 \multirow{2}{*}{Methods} & \multirow{2}{*}{Pub.’Year} &\multirow{2}{*}{Task} & \multicolumn{4}{c}{Foggy-Puddle} \\
\cline{4-7}
&&& $IoU$ & $MIoU$ & $NPA$ & $F_\beta$ \\
\cline{1-7}
FCN8s-RAU\cite{37} & ECCV'2018 &RPD & 80.93 & 90.35 & 94.97 &89.66 \\

Deeplabv3+\cite{93} & ECCV'2018 &Seg &  80.82 & 90.32 & 94.45 & 89.60 \\

CCNet\cite{ccnet} & IEEE TPAMI'2020 &Seg &  80.75 & 90.29 & 94.27 & 89.35 \\

F3Net\cite{62} & AAAI'2020 &SOD &  78.02 & 88.90 & 93.30 & 87.65 \\

U2Net\cite{71} & PR'2020 &SOD &  80.54 & 90.24 & 95.05 & 89.37 \\

Segformer-B2\cite{94} & NeurIPS'2021 &Seg &  78.33 & 89.20 & 93.35 & 88.70 \\

SWNet\cite{39} & IEEE TITS'2022 &RPD &  79.27 & 89.51 & 93.76 & 88.43 \\

GDNet\cite{gdnet} & IEEE TPAMI'2023 &SOD &  80.87 & 90.32 & 94.73 & 89.56 \\

Homofusion\cite{homofusion} & ICCV'2023 &RPD &  80.35 & 90.12 & 94.06 & 89.23 \\

SeaFormer-Base\cite{seaformer} & ICLR'2023 &Seg &  \textcolor{green}{81.06} & \textcolor{green}{90.48} & \textcolor{green}{95.18} & \textcolor{green}{89.73} \\

\textcolor{blue}{OURS} & \textcolor{blue}{- - - -} & \textcolor{blue}{RPD} & \textbf{\textcolor{blue}{81.68↑}} & \textbf{\textcolor{blue}{90.75↑}} & \textbf{\textcolor{blue}{95.86↑}} & \textbf{\textcolor{blue}{89.92↑}} \\
\hline
\end{tabular}
\label{tab:foggy}
\end{table}

\begin{table}[htbp]
\centering
\caption{Quantitative comparison data with other methods on the Night-Puddle are presented. Seg stands for segmentation, SOD represents salient object detection, and RPD signifies road ponding detection. The unit for $IoU$, $MIoU$, $PA$, $F_\beta$ is percentage ($\%$).  The first and second best results are indicated in \textbf{\textcolor{blue}{blue}}, \textcolor{green}{green}, respectively.}
\label{night_methods}
\tabcolsep=0.03cm
\renewcommand\arraystretch{1.5}
\begin{tabular}{c|c|c|cccc}
\hline
\multirow{2}{*}{Methods} & \multirow{2}{*}{Pub.’Year} &\multirow{2}{*}{Task} &  \multicolumn{4}{c}{Night-Puddle}\\
\cline{4-7}
&&& $IoU$ & $MIoU$ & $NPA$ & $F_\beta$ \\
\cline{1-7}
FCN8s-RAU\cite{37} & ECCV'2018 &RPD &  70.93 & 83.52 & 88.10 & 83.72 \\

Deeplabv3+\cite{93} & ECCV'2018 &Seg &  72.76 & 84.70 & 90.32 & 84.23 \\

CCNet\cite{ccnet} & IEEE TPAMI'2020 &Seg &  70.96 & 83.58 & 88.94 & 84.03 \\

F3Net\cite{62} & AAAI'2020 &SOD & 72.83 & 84.83 & 90.36 & 85.28 \\

U2Net\cite{71} & PR'2020 &SOD & 75.81 & 86.43 & 91.51 &86.24 \\

Segformer-B2\cite{94} & NeurIPS'2021 &Seg &  75.23 & 86.03 & 91.21 & 86.02 \\

SWNet\cite{39} & IEEE TITS'2022 &RPD & 75.72 & 86.36 & 91.44 & 86.18 \\

GDNet\cite{gdnet} & IEEE TPAMI'2023 &SOD &  74.32 & 85.77 & 90.68 & 85.64 \\

Homofusion\cite{homofusion} & ICCV'2023 &RPD &  75.47 & 86.38 & 91.37 & 86.16 \\

SeaFormer-Base\cite{seaformer} & ICLR'2023 &Seg & \textcolor{green}{75.85} & \textcolor{green}{86.52} & \textcolor{green}{91.58} & \textcolor{green}{86.28} \\

\textcolor{blue}{OURS} & \textcolor{blue}{- - - -} &\textcolor{blue}{RPD} & \textbf{\textcolor{blue}{76.91↑}} & \textbf{\textcolor{blue}{87.68↑}} & \textbf{\textcolor{blue}{92.86↑}} & \textbf{\textcolor{blue}{87.14↑}} \\
\hline
\end{tabular}
\label{tab:night_exp}
\end{table}

\begin{table}[h]
\centering
\caption{For the fundamental details of the dataset, "ONR" represents "Structured Road" and "OFR" represents "Non-Structured Road"}
\label{data}
\tabcolsep=0.25cm
\renewcommand\arraystretch{2}
\begin{tabular}{ccccc}
\hline
{Dataset Name} & {Count} & {Resolution} & {Weather} & {Road Condition}\\
\cline{1-5}
{Puddle-1000} & {985} & {640×360} & {Daytime} & {ONR/OFR} \\

{Foggy-Puddle} & {985} & {640×360} & {Foggy} & {ONR/OFR}\\

{Night-Puddle} & {500} & {640×360} & {Night} & {ONR/OFR} \\
\hline
\end{tabular}
\label{data}
\end{table}

\section{Experiment} 
\subsection{Datasets}
\subsubsection{Puddle-1000}
Our training and evaluation of AGSENet utilized the publicly available benchmark dataset, Puddle-1000 \cite{58}. It comprises two subsets that cater to various complex driving conditions: ONR and OFR. ONR includes 357 structured road ponding images, each with a resolution of 640×360, while OFR contains 628 unstructured road ponding images, also with a resolution of 640×360. We noted several instances of mislabeling and omitted labels, significantly hampering the reliability of our outcomes, as indicated by the red lines in Fig. \ref{fig:data}. As such, we employed the EISeg intelligent interactive segmentation annotation software to rectify these inaccuracies in the dataset, as depicted by the blue lines in Fig. \ref{fig:data}. To establish the robustness of our methodology in detecting road ponding across diverse road conditions, we performed experiments on ONR, OFR, and the entire Puddle-1000 dataset (MIX).

In order to enrich the dataset scenarios and promote research on road water accumulation detection in foggy weather, we constructed the Foggy-Puddle road water accumulation dataset based on Puddle-1000. Our synthetic foggy dataset is inspired by the work of Sakaridis et al.\cite{foggy}, which employs an automated process to transform clear-weather images into synthetic foggy images. Firstly, the standard atmospheric scattering model is used to simulate the effect of fog on images. This model maps the radiance of the scene to the observed radiance at the camera sensor, primarily parameterized by the depth of the scene. The specific formula is:

\begin{equation}
I(x) = R(x) \cdot t(x) + L \cdot (1 - t(x))
\end{equation}

where \( I(x) \) is the observed foggy image at pixel \( x \), \( R(x) \) is the clear-weather scene radiance, \( L \) is the atmospheric light assumed to be a global constant, and the transmission rate \( t(x) \) is determined by the distance \( \ell(x) \) from the scene to the camera and the fog attenuation coefficient \( \beta \), given by:

\begin{equation}
t(x) = \exp(-\beta \cdot \ell(x))
\end{equation}

\begin{figure*}[!t]
  \centering
  \includegraphics[width=180mm]{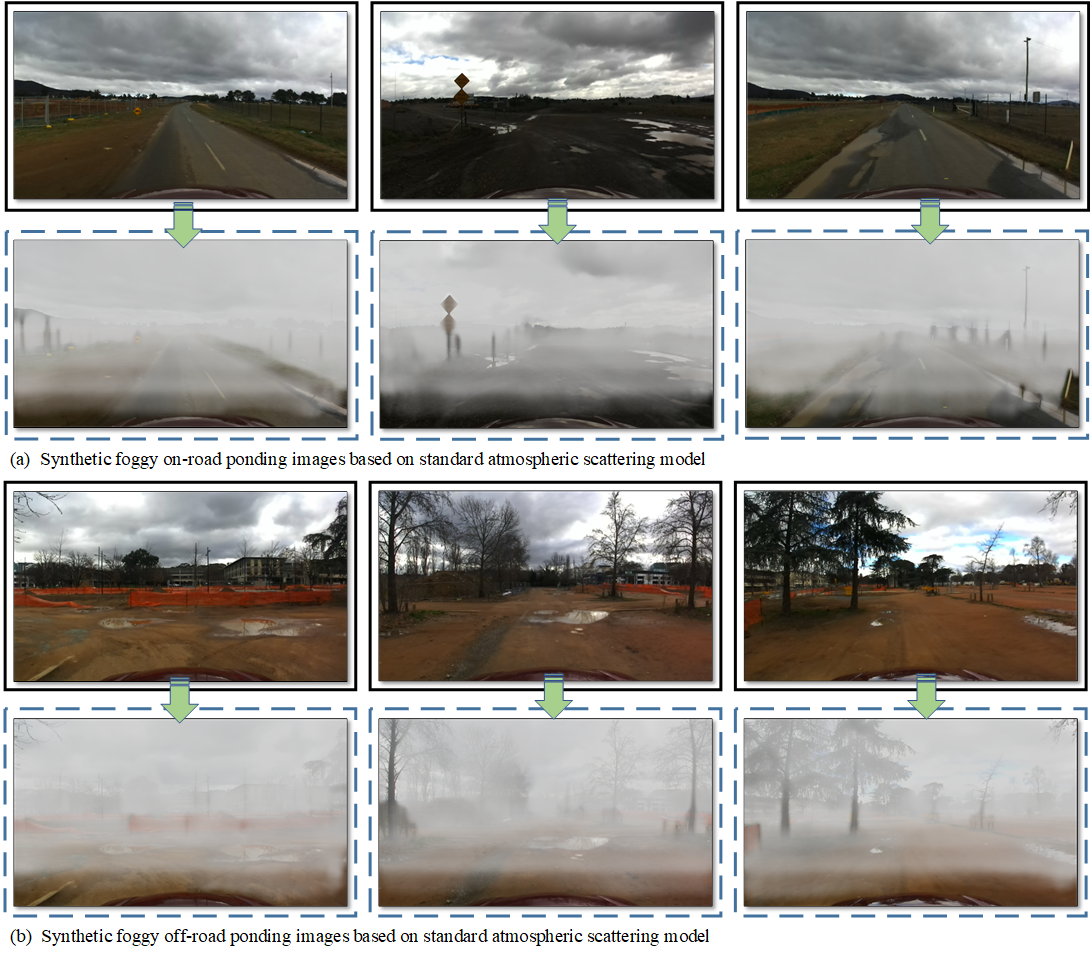}
  \caption{Representative examples of the diverse scenes encompassed within the Foggy-Puddle dataset. The first row shows structured road images extracted from the Puddle-1000 dataset, while the second row displays their corresponding synthetic foggy images generated using the atmospheric scattering model. The third row presents unstructured road images from the Puddle-1000 dataset, and the fourth row shows their corresponding synthetic foggy images created using the same atmospheric scattering model.}
  \label{foggy}
\end{figure*}

Synthesizing the fog effect requires the depth map of the scene. To obtain more accurate depth estimations, we refer to the advanced depth estimation method by Cl{\'{e}}ment Godard et al.\cite{depth}, which provides superior depth map information for the Puddle-1000 dataset compared to the method used in \cite{foggy}. Fig. \ref{foggy} shows representative examples, and Table \ref{data} provides more details.

\subsubsection{Night-Puddle}
\begin{figure}[!t]
  \centering
  \includegraphics[width=90mm]{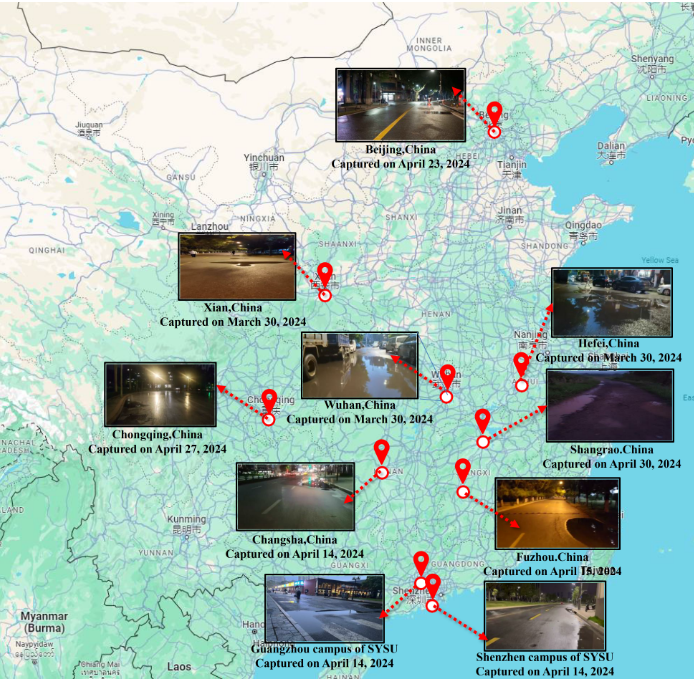}
  \caption{Comprehensive Collection of Nighttime Road Ponding Images Captured from Various Locations Across China, Including Beijing, Xi'an, Wuhan, Chongqing, Hefei, Shangrao, Changsha, Fuzhou, and the SYSU Campuses in Guangzhou and Shenzhen. Taken on Specific Dates in 2024.}
  \label{map}
\end{figure}

\begin{figure*}[!t]
  \centering
  \includegraphics[width=180mm]{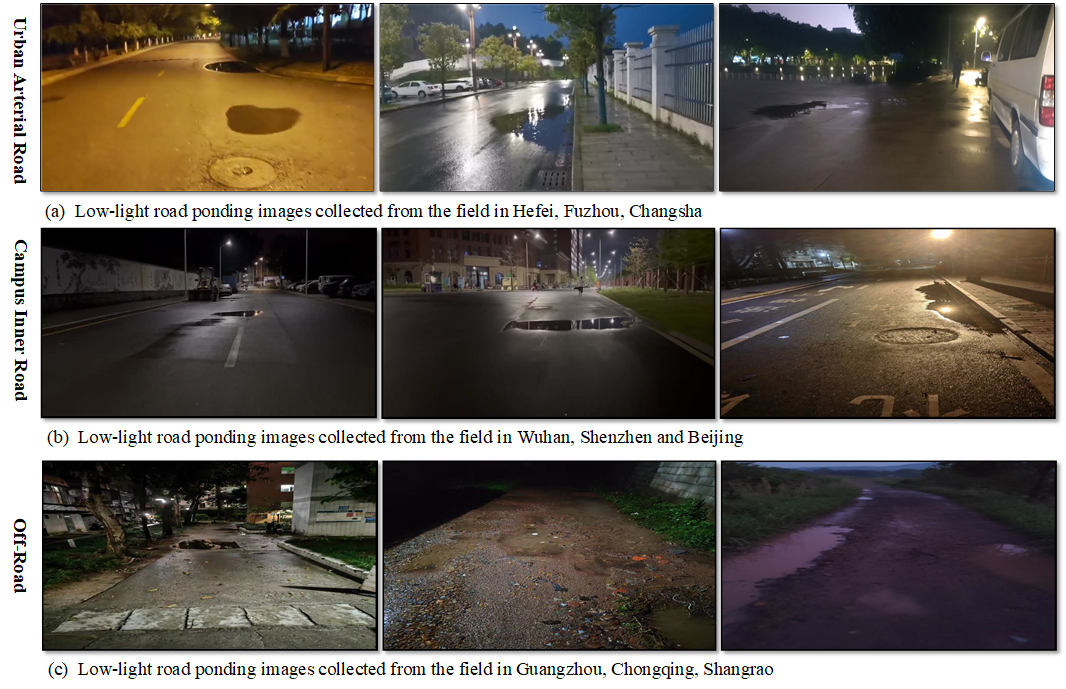}
  \caption{Representative examples of the diverse scenes encompassed within the Night-Puddle dataset. Specifically, we have collected images from images from Hefei, Fuzhou, Changsha (as shown in (a) Urban Arterial Road), Wuhan, Shenzhen, Beijing (as shown in (b) Campus Inner Road), and Guangzhou, Chongqing, Shangrao (as shown in (c) Off-Road). These locations provide a broad spectrum of urban and suburban settings, enabling us to capture the complexity and diversity of road ponding in real-world scenarios.}
  \label{NIGHT}
\end{figure*}

Nighttime is a high-risk period for traffic accidents, and road ponding further exacerbates this risk, while limited visibility makes ponding detection more challenging. To validate the stability and detection capability of our algorithm in nighttime environments, our team curated an extensive dataset named Night-Puddle, specifically tailored for nighttime and low-light conditions. We collected real nighttime ponding data from urban arterials, campuses, and unstructured roads across ten cities in China. This data collection aims to assess the algorithm's reliability under nighttime conditions. This dataset comprises 488 images of road ponding captured in authentic nighttime environments and an additional 12 images captured under low-light settings. To ensure the comprehensiveness and representativeness of the Night-Puddle dataset, we have conducted a rigorous data collection process, encompassing multiple cities with varying road conditions and environmental characteristics. Specifically, we have gathered images from Beijing, Guangzhou, Shenzhen, Xi'an, Wuhan, Chongqing, Changsha, Hefei, Shangrao, Fuzhou, as well as the Sun Yat-sen University campuses in Guangzhou and Shenzhen, such as Fig. \ref{map}. These locations offer a wide range of urban, suburban, enabling us to capture the complexity and diversity of road ponding in real-world scenarios. Fig. \ref{NIGHT} provides representative examples of the diverse scenes encompassed within the Night-dataset. Moreover, we have adhered to strict quality control measures during the annotation process to ensure the accuracy and reliability of the labels. Detailed statistics and analyses of the dataset are provided in Table \ref{data}, highlighting its unique characteristics and potential applications in advancing research on road ponding detection in challenging nighttime and low-light environments.

\subsection{Implementation Details}
Our experiments using the PyTorch toolbox and NVIDIA GeForce RTX 3090 maintained consistent training and testing strategies across ONR, OFR, and MIX. Specifically, we enhanced input images through random horizontal flipping, brightness and saturation adjustment, and resizing. All model layers were initialized with random weights in our experimentation and underwent training from scratch for 500 epochs. We used the stochastic gradient descent (SGD) optimizer to optimize the loss function and a weight decay of $5 \times 10^{-4}$. We set the batch size to 4 and the initial learning rate to 0.001. During the initial training phase, the network may struggle to capture specific class features. Therefore, we first froze the SSIE module, only unfreezing it after 50 epochs, to effectively train our network.
\subsection{Evaluation Metrics}
Our method focuses on the accurate detection and segmentation of road ponding, representing segmented areas as pixel sets. We use common semantic segmentation metrics to evaluate AGSENet's performance, including Intersection over Union (IoU), Mean Intersection over Union (MIoU), F-measure ($F_\beta$), and the Normalized Pixel Accuracy (NPA) proposed in this paper.
\begin{itemize}
\item \textbf{Intersection over Union}(IoU) is commonly used evaluation metric in the segmentation field, which is defined as:
\end{itemize}
\begin{align}
&IoU = \frac{TP}{TP + FN+ FP},
\end{align}
\begin{itemize}
\item \textbf{Mean Intersection over Union}(MIoU) measures the average ratio between the intersection and the union of the ground truth and predicted pixel regions. It is calculated as follows:
\end{itemize}
\begin{align}
&MIoU = \frac{1}{k + 1}\sum_{i=0}^k\frac{TP}{TP + FN+ FP},
\end{align}
\begin{itemize}
\item \textbf{F-measure}($F_\beta$) is a comprehensive evaluation metric that takes into account both the precision and recall of a prediction map. It is defined as follows:
\end{itemize}
\begin{align}
&Pre = \frac{TP}{TP + FP}, \\
&Rec = \frac{TP}{TP + FN}, \\
&F_\beta = \frac{2\times Pre \times Rec}{Pre + Rec},
\end{align}

\begin{itemize}
\item \textbf{Normalized Pixel Accuracy (NPA)} is designed to evaluate pixel prediction accuracy, especially for datasets with highly imbalanced distributions of positive and negative samples. In the Puddle-1000 dataset, for example, negative samples account for 98.5\% of the total. This imbalance makes traditional pixel-level accuracy metrics ineffective for evaluating network performance. To address this issue, we propose NPA, which uses true positive rate (TPR), false positive rate (FPR), true negative rate (TNR), and false negative rate (FNR) instead of raw counts. This approach allows for a more effective evaluation of datasets with a skewed positive-to-negative ratio. The formula for NPA is as follows:
\end{itemize}
{\begin{align}
&TPR = \frac{TP}{TP + FN}, \\
&TNR = \frac{TN}{TN + FP}, \\
&FNR = \frac{FN}{TP + FN}, \\
&FPR = \frac{FP}{TN + FP}, 
\end{align}
\begin{align}
&NPA = \frac{TPR + TNR}{TPR + TNR + FPR + FNR} \nonumber \\
&\phantom{NPA} = \frac{TPR + TNR}{2},
\end{align}

\indent\setlength{\parindent}{2em}Where TP represents the count of true positive pixels; TN represents the count of true negative pixels; FP represents the count of false positive pixels; FN represents the count of false negative pixels; TPR represents the true positive rate; TNR represents the true negative rate; FPR represents the false positive rate; FNR represents the false negative rate; k is the count of classes, including one background class.}
\begin{figure*}
    \centering
    \includegraphics[scale=0.52]{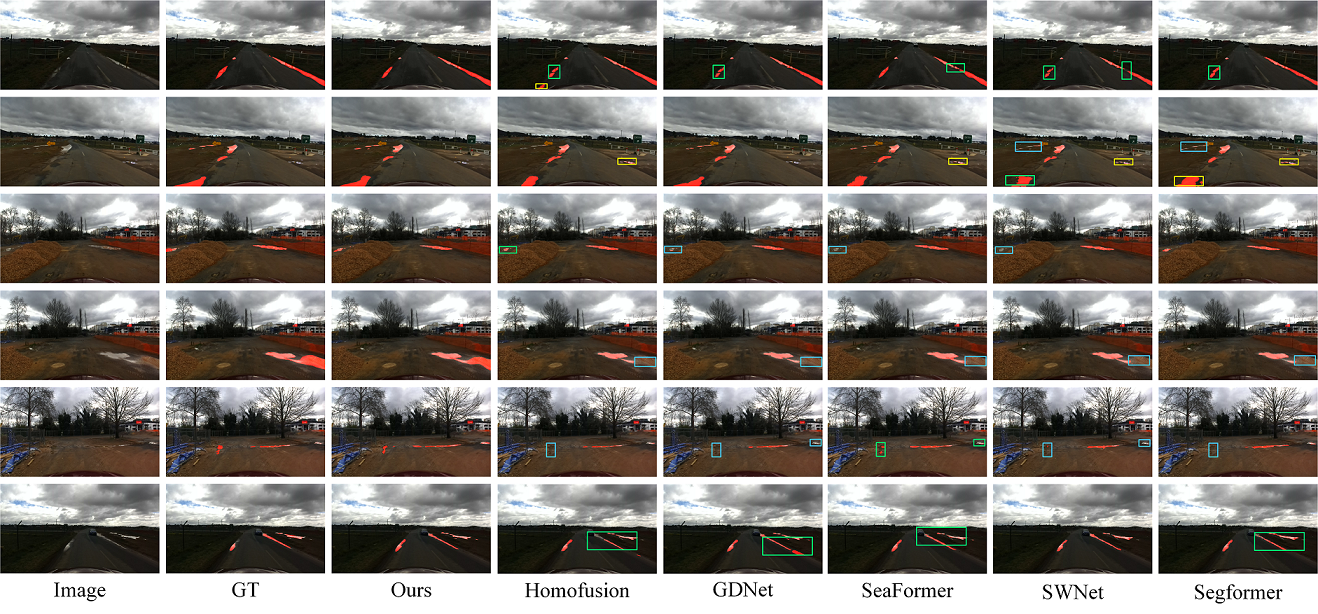}
    \caption{In the Puddle-1000(ONR), Puddle-1000(OFR), and Puddle-1000(MIX) datasets, the prediction results of AGSENet(Ours), Homofusion, GDNet, SeaFormer, SWNet, and Segformer are presented. From top to bottom, they correspond to two selected images from each dataset. Yellow boxes indicate false positives, blue boxes indicate missed detections, and green boxes indicate discontinuous detections and rough edges.}
    \label{fig:ks1}
\end{figure*}
\begin{figure*}
    \centering
    \includegraphics[scale=0.52]{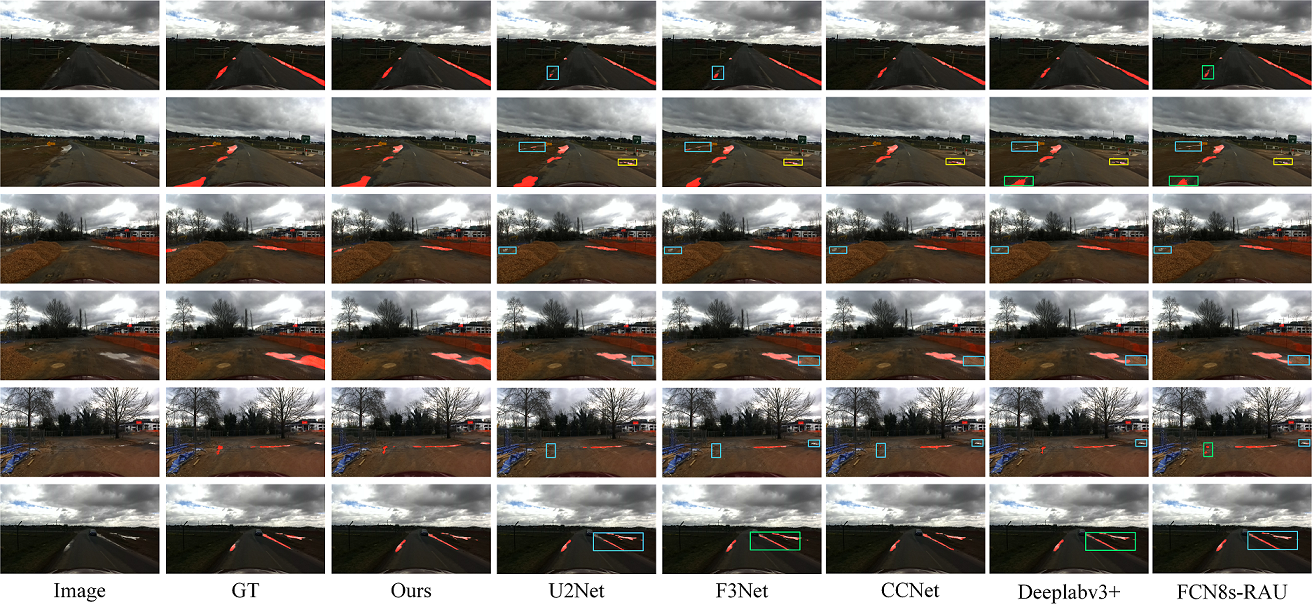}
    \caption{In the Puddle-1000(ONR), Puddle-1000(OFR), and Puddle-1000(MIX) datasets, the prediction results of AGSENet(Ours), U2Net, F3Net, CCNet, Deeplabv3+, and FCN8s-RAU are presented. From top to bottom, they correspond to two selected images from each dataset. Yellow boxes indicate false positives, blue boxes indicate missed detections, and green boxes indicate discontinuous detections and rough edges.}
    \label{fig:ks2}
\end{figure*}
\begin{figure*}
    \centering
    \includegraphics[scale=0.32]{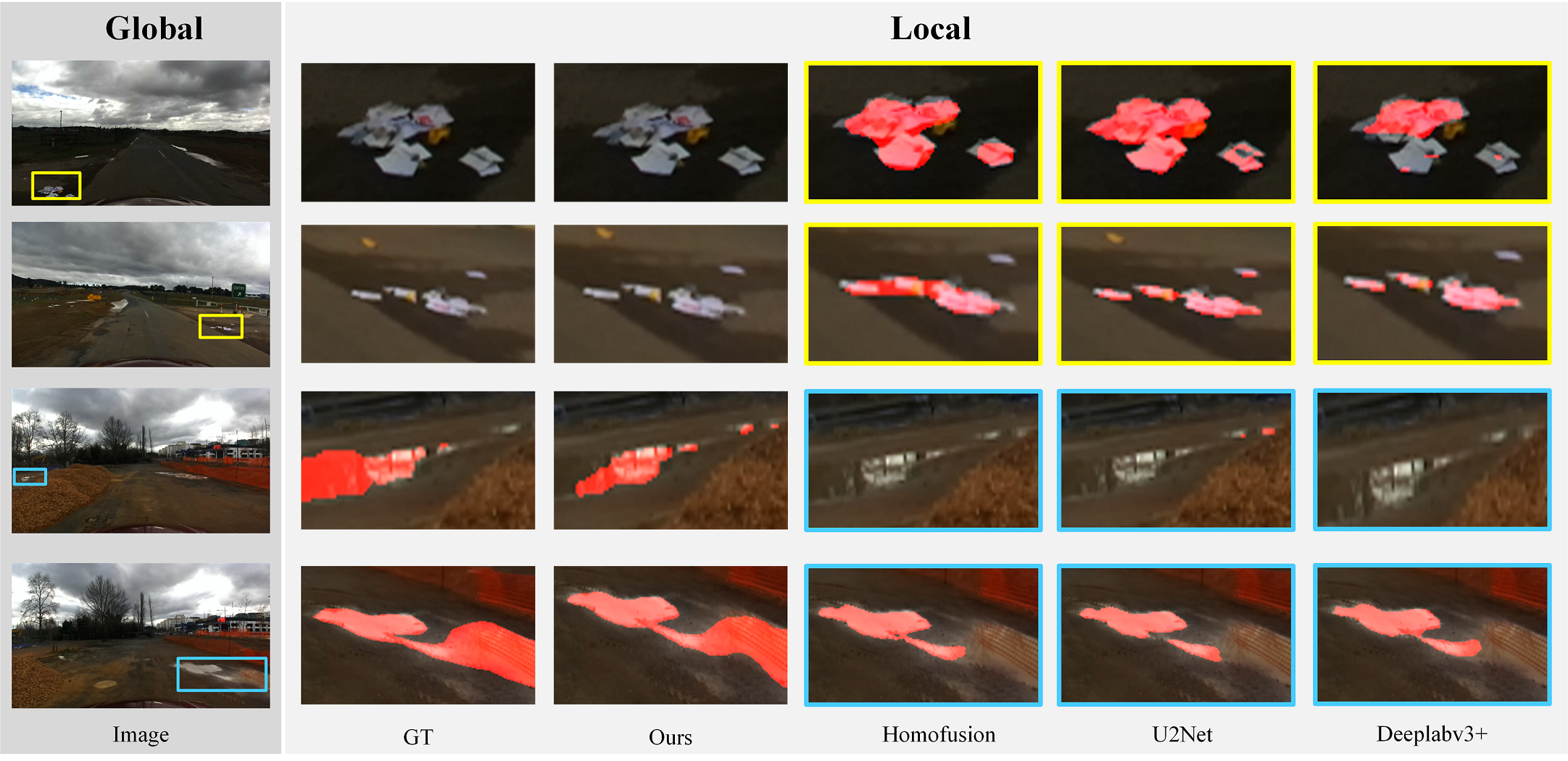}
    \caption{Visualized images of locally enlarged prediction results from AGSENet (our method), Homofusion, U2Net, and Deeplabv3+. Yellow boxes indicate false positives, while blue boxes indicate missed detections.}
    \label{detail}
\end{figure*}

\subsection{Comparison With the State-of-the-Art Methods}
To evaluate AGSENet's performance, we compared it with ten primary methods based on their classical architectures, recent publications, and latest advancements in road ponding detection and related fields. These methods include DeepLabv3+ \cite{83}, Segformer \cite{84}, CCNet, and Seaformer-Base for semantic segmentation; U2Net \cite{79}, F3Net \cite{20}, and GDNet for salient object detection (SOD); and SWNet \cite{61}, FCN8s-RAU \cite{58}, and Homofusion for road ponding detection (RPD). For fairness, we used publicly accessible code and conducted all experiments in a unified environment, evaluating all prediction maps using the same codebase.

\subsubsection{Puddle-1000 (ONR)}

Table \ref{ONR} (columns 4-7) presents the quantitative results of our approach compared to other methods on the Puddle-1000 (ONR) dataset. AGSENet outperforms all other methods across four standard evaluation metrics. Specifically, AGSENet achieves a 3.19\% improvement in IoU compared to FCN8s-RAU, a 2.99\% improvement over SWNet, and a 2.96\% improvement over Homofusion. Furthermore, our approach shows up to an 8.31\% IoU improvement compared to other representative methods.

Figures \ref{fig:ks1} and \ref{fig:ks2} illustrate the detection results of all compared methods on the Puddle-1000 (ONR) dataset. The first two rows show prediction results, where the green boxes indicate our method's more continuous ponding edge detection. The blue and yellow boxes highlight AGSENet's robustness and noise immunity. This is primarily due to our proposed SSIE module, which effectively suppresses background noise and enhances edge detail information. Additionally, AGSENet can detect road puddles in distant areas, significantly enhancing the early warning capabilities of road safety monitoring systems.

\subsubsection{Puddle-1000(OFR)}

Table \ref{ONR} (columns 8-11) shows the performance results of AGSENet on the Puddle-1000 (OFR) dataset, where it surpasses all other methods. AGSENet achieves a 1.52\% improvement in IoU compared to FCN8s-RAU, a 2.73\% improvement over SWNet, and a 1.13\% improvement over Homofusion. Additionally, AGSENet enhances IoU by up to 2.66\% compared to other methods.


The middle two rows of Figures \ref{fig:ks1} and \ref{fig:ks2} illustrate AGSENet's robust detection results. The blue boxes indicate that existing methods often miss detections due to reflections on water surfaces or subtle road ponding. In contrast, AGSENet excels in detecting such road ponding. This effectiveness is primarily due to the CSIF module's ability to capture long-range similarity information in the channel dimension, enabling better identification of salient features induced by reflections.

\begin{figure*}
    \centering
    \includegraphics[scale=0.5]{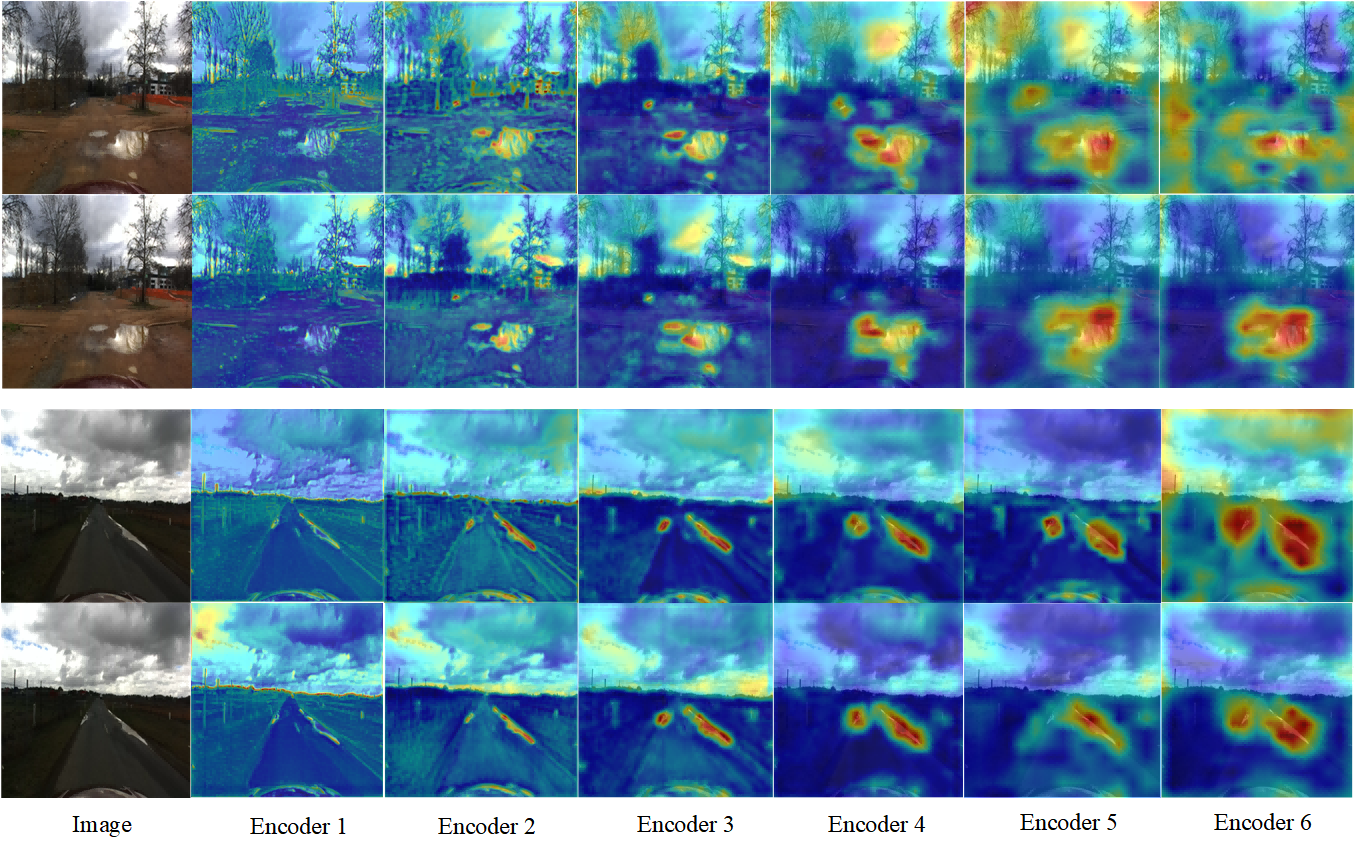}
    \caption{The heatmaps of different stages of the Encoder are presented in sequence from top to bottom as follows: without the addition of CSIF, with the addition of CSIF, without the addition of CSIF, and with the addition of CSIF.}
    \label{fig:out-csie}
\end{figure*}

\subsubsection{Puddle-1000(MIX)}

To evaluate AGSENet's robustness, we conducted experiments on the Puddle-1000 (MIX) dataset. The results in Table \ref{ONR} (columns 12-15) show that AGSENet outperforms all compared methods. Specifically, AGSENet improves IoU by 2.23\% compared to FCN8s-RAU, 4.25\% compared to SWNet, and 2.19\% compared to Homofusion. Additionally, AGSENet achieves up to a 5.56\% IoU improvement over other methods.

The last two rows of Figures \ref{fig:ks1} and \ref{fig:ks2} illustrate AGSENet's superior performance on unstructured roads, effectively addressing issues like discontinuity and missed detections for ponding water that closely resembles the road surface in color. Moreover, AGSENet demonstrates robust performance in detecting distant or subtle road ponding areas.

To clearly demonstrate the effectiveness of our method, we provide visualized images of locally enlarged typical scenarios, comparing them with the latest ponding detection network, Homofusion; the classic salient object detection network, U2Net; and the typical semantic segmentation network, Deeplabv3+. As shown in Figure \ref{detail}, our model not only avoids false detections and missed detections but also adapts robustly to annotation defects.

\subsubsection{Foggy-Puddle}
Detecting road ponding in foggy weather is crucial due to significantly reduced visibility and increased risk of accidents. To evaluate AGSENet's performance under such conditions, we conducted experiments on our self-constructed Foggy-Puddle dataset. As shown in Table \ref{tab:foggy}, AGSENet outperformed all comparison methods. Specifically, AGSENet achieved a 0.75\% improvement in IoU compared to FCN8s-RAU, 2.41\% over SWNet, and 1.33\% over Homofusion. Additionally, AGSENet improved IoU by 0.81\% compared to GDNet, 3.66\% compared to F3Net, and 0.62\% compared to Seaformer. Therefore, our method demonstrates robust road ponding detection capabilities under foggy weather conditions, highlighting its potential to enhance traffic safety in adverse weather.

\subsubsection{Night-Puddle}
Road ponding detection at night is also critical due to limited visibility and higher likelihood of accidents. To evaluate AGSENet's generalization capability in low-light conditions, we conducted experiments on our self-constructed Night-Puddle dataset. As shown in Table \ref{tab:night_exp}, AGSENet outperformed all compared methods. Specifically, AGSENet achieved a 5.98\% improvement in IoU over FCN8s-RAU, 1.19\% over SWNet, and 1.44\% over Homofusion. Additionally, AGSENet improved IoU by 2.59\% compared to GDNet, 4.08\% compared to F3Net, and 1.06\% compared to Seaformer.

These results indicate that despite the Night-Puddle dataset containing only 500 samples, AGSENet efficiently extracts salient features of road ponding under limited dataset conditions and stably detects road ponding in low-light and nighttime scenes. Our experiments confirm that under normal daytime conditions, AGSENet effectively utilizes the reflection characteristics of road ponding to capture long-range similarity information while refining the edges. Additionally, AGSENet achieves state-of-the-art (SOTA) performance under both foggy weather and low-light nighttime conditions. In summary, AGSENet effectively identifies and robustly detects salient features of road ponding in complex driving scenarios, making it a valuable tool for improving road safety in challenging weather and lighting conditions.

\begin{table}[htb]
\centering
\caption{The ablation experiment results indicate that each component in AGSENet contributes to the overall performance. “B” denotes base network without CSIF module and SSIE module. “SCIP”, “CSII” are Spatial Context Information Perception, and Channel Significant Information Interaction in CSIF module, respectively. “NR”, “ER” are Noise Rejection, and Edge Refinement in SSIE module, respectively. The unit for all experimental results is percentage ($\%$).}
\label{AS}
\tabcolsep=0.22cm
\renewcommand\arraystretch{1.3}
\begin{tabular}
{cl|cccc}
\hline
\multicolumn{2}{c|}{\multirow{2}{*}{Methods}}&\multicolumn{4}{c}{Puddle-1000(MIX)} \\
\cline{3-6}
& & $IoU$ & $MIoU$ & $NPA$ & $F_\beta$ \\
\cline{1-6}
$(a)$ &B &82.97 &91.39 &94.88 &90.69 \\ 
$(b)$ &B+SCIP &84.72 &92.28 &95.85 &91.73 \\
$(c)$ &B+CSII &84.91 &92.38 &95.91 &91.84 \\
$(d)$ &B+CSIF &85.31 &92.58 &96.34 &92.07 \\
\hline
$(e)$ &B+NR &84.88 &92.36 &95.77 &91.82 \\
$(f)$ &B+ER &84.71 &92.28 &95.85 &91.72 \\
$(g)$ &B+SSIE &84.92 &92.38 &96.17 &91.84 \\
\hline
$\textcolor{blue}{(h)}$ &\textcolor{blue}{B+CSIF+SSIE(OURS)} &\textbf{\textcolor{blue}{85.84↑}} &\textbf{\textcolor{blue}{92.85↑}} &\textbf{\textcolor{blue}{96.68↑}} &\textbf{\textcolor{blue}{92.38↑}} \\
\hline
\end{tabular}
\label{AS}
\end{table}
\begin{figure*}
    \centering
    \includegraphics[scale=0.5]{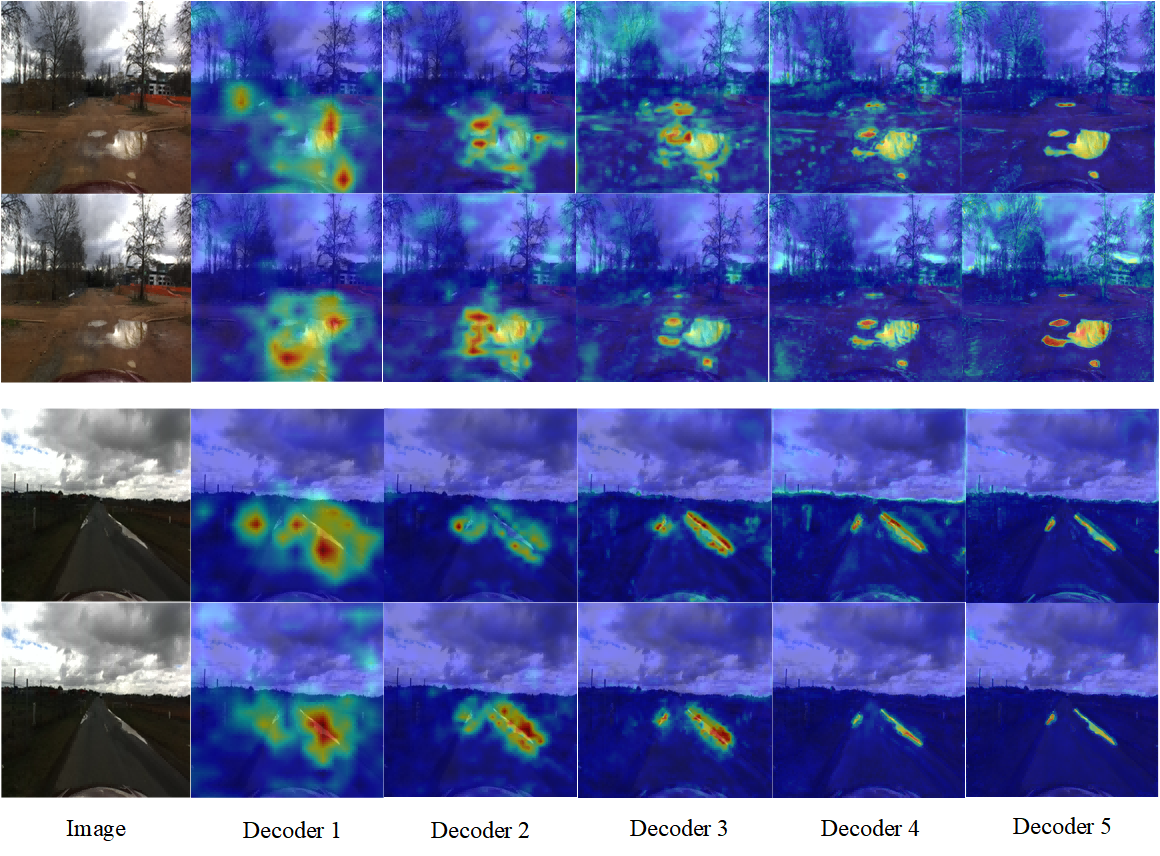}
    \caption{The heatmaps of different stages of the Decoder are presented in sequence from top to bottom as follows: without the addition of SSIE, with the addition of SSIE, without the addition of SSIE, and with the addition of SSIE.}
    \label{fig:out-ssif}
\end{figure*}
\subsection{Ablation Study} 
To ascertain the effectiveness of the CSIF and SSIE modules, we conducted an ablation study on the Puddle-1000 (MIX) dataset under consistent conditions. The results are compiled in Table \ref{AS}, Fig. \ref{fig:out-csie}, and Fig. \ref{fig:out-ssif}. These findings offer valuable insights into the performance contributions of these two critical components.

\textbf{Effectiveness of the CSIF Module:} The CSIF module integrates two subcomponents, Spatial Context Information Perception (SCIP) and Channel Significant Information Interaction (CSII), tasked with augmenting salient features. We initially trained a base network (``B") devoid of both the CSIF and SSIE modules. Subsequently, SCIP and CSII were incrementally introduced into the CSIF module. Analysis of the data in Table \ref{AS} allows us to make several key observations: (1) The CSIF block enhances the channel dimension's long-range similarity information, enabling a more effective capture of salient features introduced by reflection phenomena, thereby significantly improving segmentation performance (Comparison $(d)$ versus $(a)$). (2) SCIP aids in extracting richer contextual information in the spatial dimension, which helps in the identification of salient features (Comparison $(b)$ versus $(a)$). This also suggests that enhancing pixel representation capability in the spatial dimension before the interaction of long-range information in the channel dimension can compensate for information loss caused by spatial dimension collapse (Comparison $(d)$ versus $(c)$). (3) CSII enhances long-range similar information in the channel dimension (Comparison $(c)$ versus $(a)$) and further concentrates on salient similarity features, leading to improved segmentation performance (Comparison $(d)$ versus $(b)$). Fig. \ref{fig:out-csie} shows that using the CSIF module enables the network to focus more on road ponding's salient features while reducing interference from background features. This suggests that the CSIF module effectively captures global similarity information and enhances salient features.

\setlength{\parskip}{0pt}
\textbf{Effectiveness of the SSIE Module:} To authenticate the effectiveness of the SSIE module, we progressively introduced Edge Refinement (ER) and Noise Rejection (NR) components into the SSIE module, beginning with the base model. According to Table \ref{AS}, the following observations can be made: (1) The SSIE module bridges the feature disparity between road ponding features at varying levels, enhancing the salient information in the feature space and thereby improving segmentation performance ($(g)$ outperforms $(a)$); (2) NR accentuates common features across different levels during fusion, eliminates background noise from low-level features, and mitigates the influence of road ponding reflection phenomena ($(e)$ outperforms $(a)$); (3) ER explores feature discrepancies, refines the edge position data in high-level features, and hence, delivers superior performance ($(f)$ outperforms $(a)$). As illustrated in Fig. \ref{fig:out-ssif}, incorporating the SSIE module in the network bolsters the refinement of road ponding edges, making the network more attuned to salient road ponding features. This signifies that the SSIE module effectively eliminates blurred features, emphasizes the salient features of road ponding, and proficiently addresses the issue of feature edge information loss during fusion. The combined potency of the CSIF and SSIE modules (i.e., $(h)$) endows our methodology with robust road ponding detection capabilities.



In addition, to validate the effectiveness of the SSIE module's fusion strategy, we compared it with two other commonly used fusion strategies: addition and multiplication fusion. The experimental results are shown in Table \ref{tab:fusion_methods}. The addition (b) and multiplication (c) fusion strategies perform similarly. However, due to the characteristic gap between features at different levels and the redundant information in the fusion results, these simple fusion strategies limit the accuracy of ponding segmentation tasks. By highlighting the common features between different levels and introducing the exploration of feature differences (d), better performance can be achieved.

\begin{table}[htb]
\centering
\caption{Comparison of Different Fusion Methods. The ablation experiment results indicate that each component in AGSENet contributes to the overall performance. “B” denotes base network without CSIF module and SSIE module. “SCIP”, “CSII” are Spatial Context Information Perception, and Channel Significant Information Interaction in CSIF module, respectively. The unit for all experimental results is percentage ($\%$).}
\label{AS}
\tabcolsep=0.15cm
\renewcommand\arraystretch{1.3}
\begin{tabular}
{cl|cccc}
\hline
\multicolumn{2}{c|}{\multirow{2}{*}{Methods}}&\multicolumn{4}{c}{Puddle-1000(MIX)} \\
\cline{3-6}
& & $IoU$ & $MIoU$ & $NPA$ & $F_\beta$ \\
\cline{1-6}
$(a)$ &B &82.97 &91.39 &94.88 &90.69 \\ 
$(b)$ &B+CSIF+Addition Fusion &83.73 &91.85 &95.72 &91.33 \\
$(c)$ &B+CSIF+Multiplication Fusion &83.51 &91.67 &95.46 &91.04 \\
$\textcolor{blue}{(d)}$ &\textcolor{blue}{B+CSIF+SSIE(OURS)} &\textbf{\textcolor{blue}{85.84↑}} &\textbf{\textcolor{blue}{92.85↑}} &\textbf{\textcolor{blue}{96.68↑}} &\textbf{\textcolor{blue}{92.38↑}} \\
\hline
\end{tabular}
\label{tab:fusion_methods}
\end{table}

\subsection{Computational Cost}
\begin{figure}
    \centering
    \includegraphics[scale=0.23]{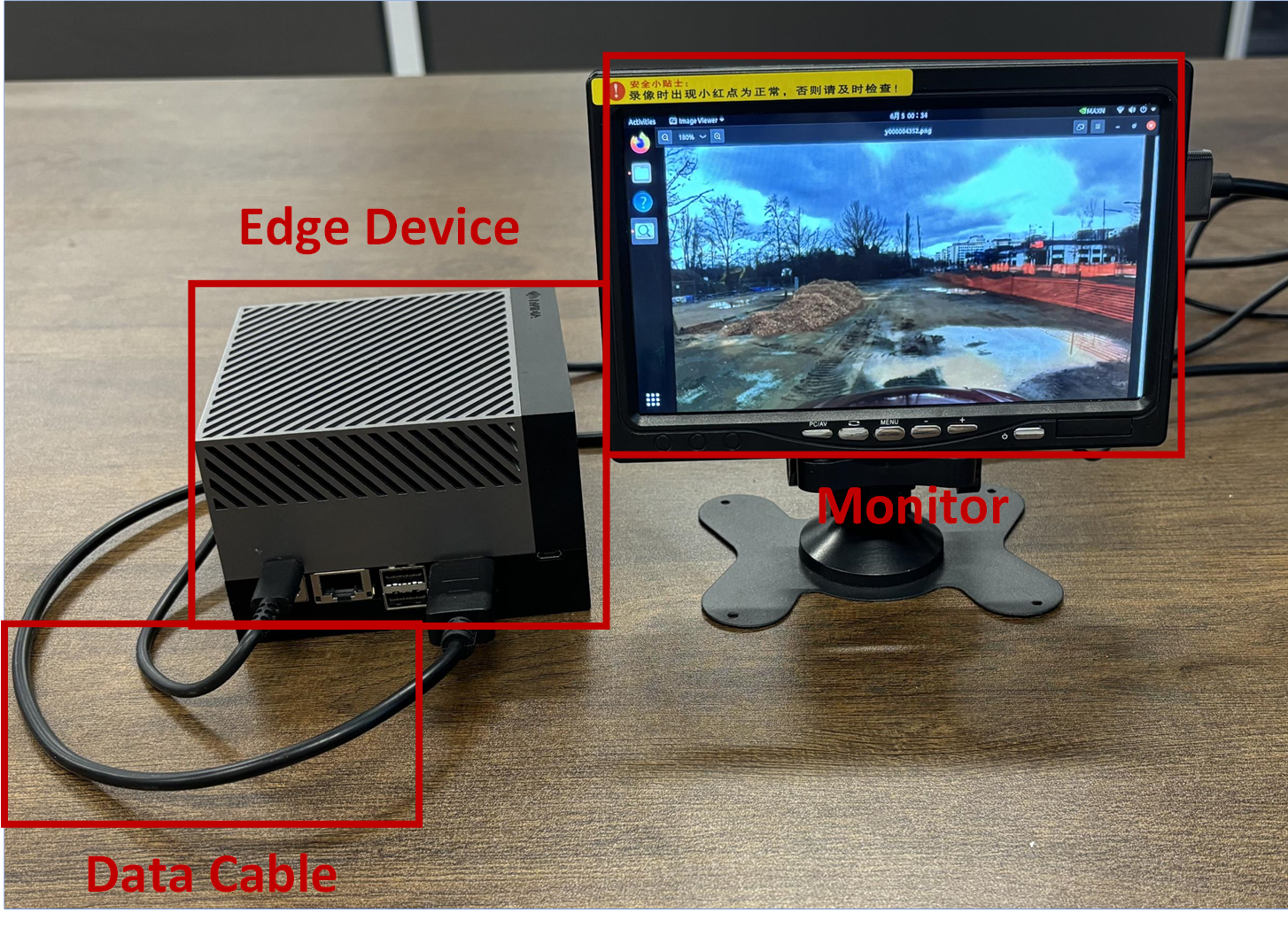}
    \caption{Edge Computing Device NVIDIA Jetson AGX Orin.}
    \label{device}
\end{figure}

To better evaluate the real-time performance of the AGSENet model, we utilized the NVIDIA Jetson AGX Orin mobile edge computing device, shown in Fig. \ref{device}, as the testing platform for the AGSENet road ponding detection algorithm.

\begin{table}[ht]
\centering
\caption{Parameters for  NVIDIA Jetson AGX Orin and NVIDIA GeForce RTX 3060}
\label{AS}
\tabcolsep=0.1cm
\renewcommand\arraystretch{1.6}
\begin{tabular}
{c|c|c}
\hline
Parameter & \parbox{2cm}{\centering {NVIDIA Jetson}\\AGX Orin} & \parbox{2.5cm}{\centering {NVIDIA GeForce}\\ RTX 3060} \\
\hline
CUDA Cores                   & 2048             & 3584             \\
Boost Clock (GHz)            & 1.30             & 1.78             \\
Standard Memory Configuration& 32GB LPDDR5      & 12GB GDDR6       \\
Compute Capability (FP32)    & 3.85 TFLOPs    & 12.72 TFLOPs    \\
Memory Interface Width       & 256-bit          & 192-bit          \\
TDP (W)                      & 15               & 170              \\
\hline
\end{tabular}
\label{tab:device}
\end{table}

AGSENet was trained on an NVIDIA GeForce RTX 3060, and real-time testing was conducted on the NVIDIA Jetson AGX Orin. The specific parameters of these devices are shown in Table \ref{tab:device}. As indicated in Table \ref{fps}, while AGSENet introduces a marginal decrease in computational speed compared to the baseline U2Net model (22.74 FPS vs. 25.51 FPS), it significantly outperforms in terms of parameter efficiency. AGSENet’s parameter size is only 2.05M, slightly higher than U2Net’s 1.77M but considerably lower than most other methods, such as FCN8s-RAU with 141.68M and GDNet with 45.11M.

Notably, while FCN8s-RAU has been a significant milestone in the field, its computational complexity exceeds ours by over tenfold, making it less practical for real-time applications. Additionally, despite achieving a higher FPS, models like SWNet and SeaFormer-Base have significantly larger parameter sizes (21.81M and 8.60M, respectively), which can be limiting for deployment in resource-constrained environments. AGSENet offers a balanced approach with competitive detection performance, a much smaller parameter size (2.05M), and an acceptable FPS (22.74), demonstrating substantial advancements over older models while maintaining practical efficiency for real-time applications.

Overall, AGSENet excels in maintaining a low parameter count while achieving competitive performance. Despite not having the highest FPS, its efficiency in terms of model size and computational complexity makes it a viable option for early-warning systems in vehicles, where both computational resources and speed need to be balanced effectively.

\begin{table}[htbp]
\begin{threeparttable}
\centering
\caption{Comparison of Model Parameters and FPS;  The first and second best results are indicated in \textbf{\textcolor{blue}{blue}}, \textcolor{green}{green}, respectively.}
\label{fps}
\tabcolsep=0.12cm
\renewcommand\arraystretch{1.5}
\begin{tabular}{c|c|c|c|c}
\hline
\multirow{2}{*}{Methods} & \multirow{2}{*}{Pub.’Year} &\multirow{2}{*}{Task} &  \multicolumn{2}{c}{Puddle-1000}\\
\cline{4-5}
&&& $Params(M)$ & $FPS$ \\
\cline{1-5}
FCN8s-RAU\cite{37} & ECCV'2018 &RPD & 141.68 & 22.15 \\

Deeplabv3+\cite{93} & ECCV'2018 &Seg & 59.34  & 28.36 \\

CCNet\cite{ccnet} & IEEE TPAMI'2020 &Seg & 28.79  & 22.33 \\

F3Net\cite{62} & AAAI'2020 &SOD & 25.54  &31.97 \\

U2Net\cite{71} & PR'2020 &SOD & \textcolor{green}{1.77}   & 25.51 \\

Segformer-B2\cite{94} & NeurIPS'2021 &Seg & 27.45  & 24.71 \\

SWNet\cite{39} & IEEE TITS'2022 &RPD  & 21.81  & \textcolor{green}{38.83} \\

GDNet\cite{gdnet} & IEEE TPAMI'2023 &SOD & 45.11 & 29.66 \\

Homofusion\cite{homofusion} & ICCV'2023 &RPD & \textbf{\textcolor{blue}{1.24}}   & 25.41 \\

SeaFormer-Base\cite{seaformer} & ICLR'2023 &Seg & 8.60   & \textbf{\textcolor{blue}{42.68}} \\

OURS &- - - - &RPD & 2.05  & 22.74 \\

\hline
\end{tabular}
\label{fps}
\begin{tablenotes}
\item \textbf{Note:} Our model, despite small parameters, lacks optimal speed due to its self-attention-like mechanism, which captures complex features through channel interactions but has high computational complexity, affecting inference speed for large inputs.
\end{tablenotes}
\end{threeparttable}
\end{table}

\section{Conclusion}
Detecting road ponding is crucial for traffic safety, particularly in foggy and low-light nighttime conditions, where reliable detection can significantly reduce the risk of accidents. Current methods often fail to meet the necessary accuracy and efficiency for effective road monitoring systems. In this paper, we present AGSENet, a novel road ponding detection method inspired by salient object detection and utilizing an encoder-decoder structure. The encoder integrates the Channel Saliency Information Focus (CSIF) module, which uses a self-attention mechanism to fuse spatial and channel information, enhancing salient channel features. The decoder incorporates the Spatial Saliency Information Exploration (SSIE) module, which leverages spatial self-attention to explore complex ponding background information and bridge feature disparities across different levels.

We conducted extensive experiments on the re-annotated Puddle-1000 dataset, demonstrating the advanced performance of our method. To validate AGSENet's effectiveness in foggy and low-light nighttime conditions, we constructed two datasets: Foggy-Puddle, a generated foggy road ponding dataset, and Night-Puddle, a real-scene nighttime road ponding dataset. Our experiments on these datasets confirmed the reliability, effectiveness, and robustness of AGSENet in challenging conditions. Additionally, we addressed the issue of sample imbalance in the Puddle-1000 dataset by proposing the NPA metric, which effectively measures the accuracy of road ponding detection.

Future work will focus on building a richer low-light nighttime dataset and exploring tailored road ponding detection methods for such scenarios.

\section*{Acknowledgements}
This work was partially supported by the National Natural Science Foundation of China (Grant Nos.52172350, 51775565), Guangdong Natural Science Foundation (Nos. 2021A1515011794, 2021B1515120032, 2022B1515120072), Guangzhou Science and Technology Plan Project (No. 202206030005), Shenzhen Key Science and Technology Program (No. JSGG20210802153412036).  

We thank Miss. Dan Yang for her valuable comments. \textbf{\textcolor{blue}{We would like to release the source codes, the generated foggy weather dataset, and the constructed nighttime dataset used in this paper.}}

\bibliographystyle{IEEEtran}
\bibliography{IEEEabrv,reference}

\begin{IEEEbiography}[{\includegraphics[width=1in,height=1.25in,clip,keepaspectratio]{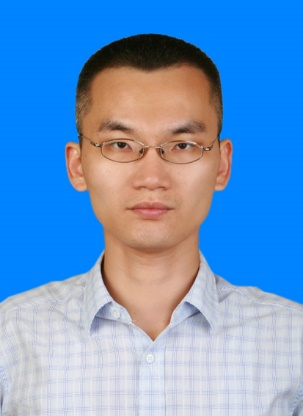}}]{Ronghui Zhang} received a B.Sc. (Eng.) from the Department of Automation Science and Electrical Engineering, Hebei University, Baoding, China, in 2003, an M.S. degree in Vehicle Application Engineering from Jilin University, Changchun, China, in 2006, and a Ph.D. (Eng.) in Mechanical \& Electrical Engineering from Changchun Institute of Optics, Fine Mechanics and Physics, the Chinese Academy of Sciences, Changchun, China, in 2009. After finishing his post-doctoral research work at INRIA, Paris, France, in February 2011, he is currently an Associate Professor with the Research Center of Intelligent Transportation Systems, School of intelligent systems engineering, Sun Yat-sen University, Guangzhou, Guangdong 510275, P.R.China. His current research interests include computer vision, intelligent control and ITS. 
\end{IEEEbiography}

\vspace{1ex}
\begin{IEEEbiography}[{\includegraphics[width=1in,height=1.25in,clip,keepaspectratio]{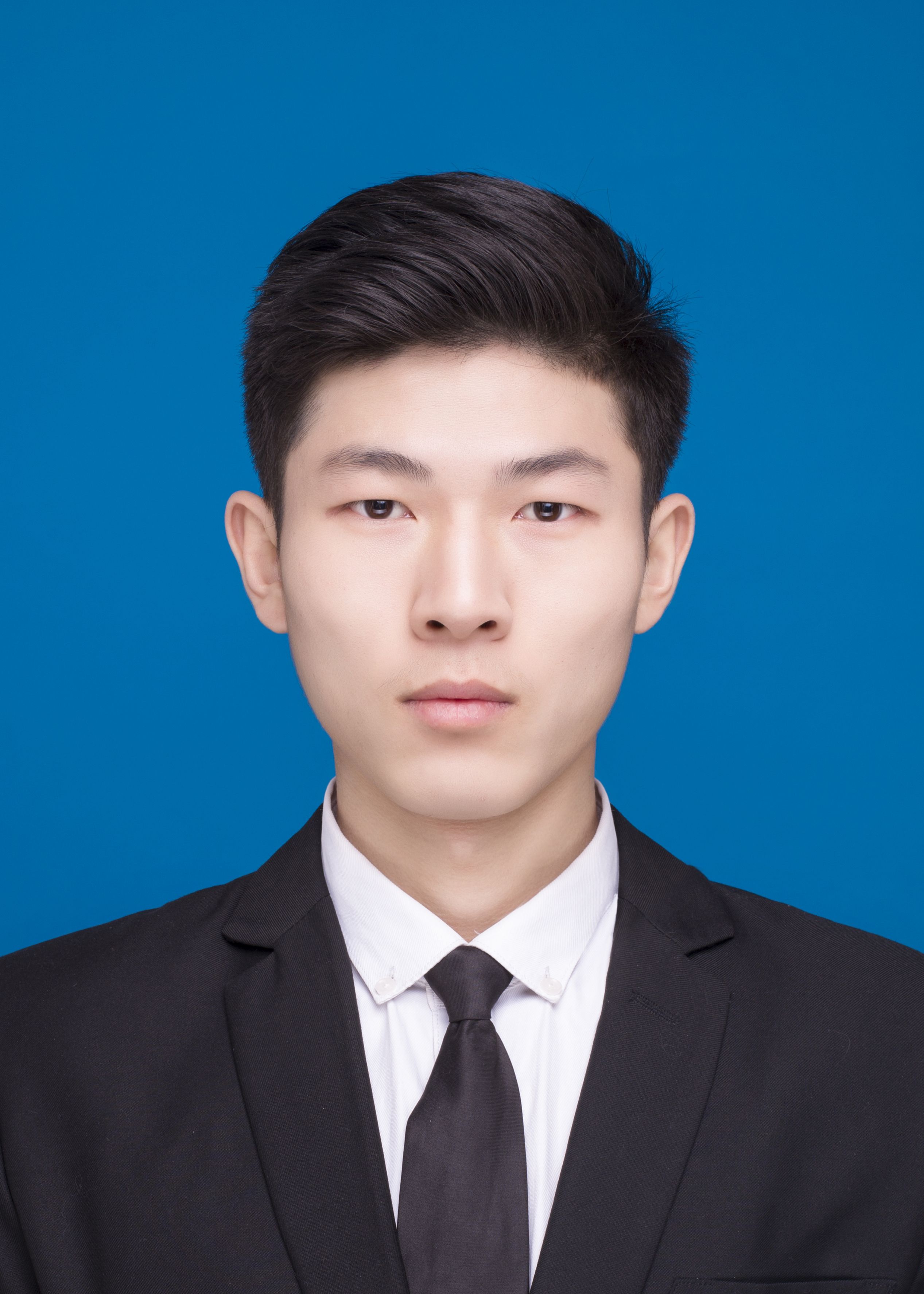}}]{Shangyu Yang} received the B.S. degree from the School of Transportation and Logistics Engineering, Shandong Jiaotong University, Jinan, China, in 2021. He is currently pursuing the M.S. degree with the School of Intelligent Systems Engineering, Sun Yat-sen University. His research interests focus on computer vision and intelligent transportation. 
\end{IEEEbiography}

\vspace{1ex}
\begin{IEEEbiography}[{\includegraphics[width=1in,height=1.25in,clip,keepaspectratio]{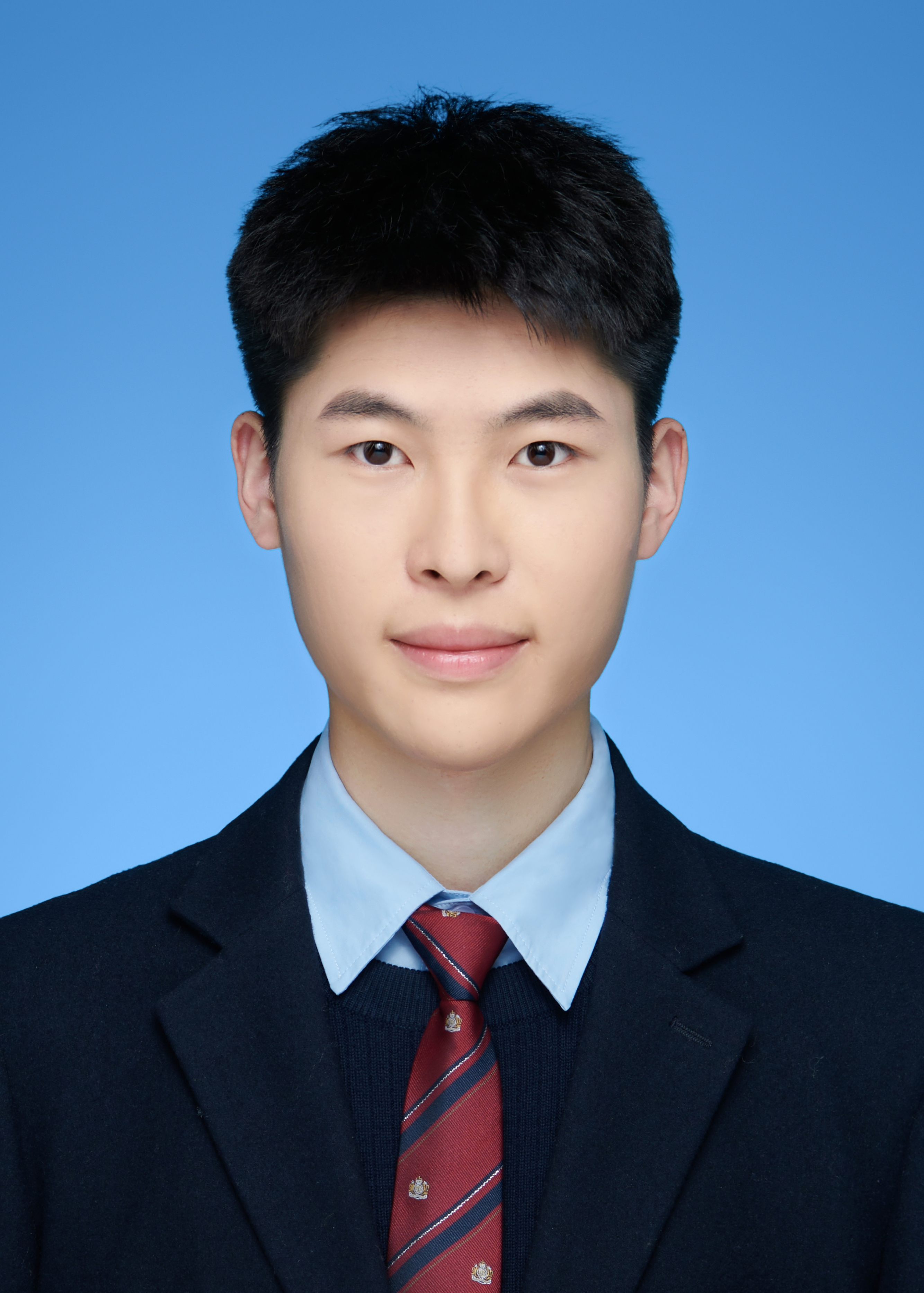}}]{Dakang Lyu} received the B.S. degree from the School of Automotive, Chang'an University, Xi'an, China, in 2023. He is currently pursuing the M.S. degree with the School of Intelligent Systems Engineering, Sun Yat-sen University. His research interests focus on computer vision and autonomous driving. 
\end{IEEEbiography}

\vspace{1ex}
\begin{IEEEbiography}[{\includegraphics[width=1in,height=1.25in,clip,keepaspectratio]{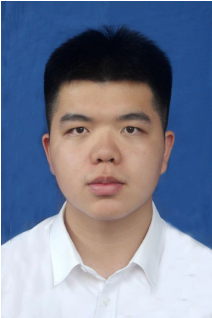}}]{Zihan Wang} received his B.S. degree in Trafﬁc Engineering from Chang’an University, Xi'an,
China, in 2021. He is currently working towards his master’s degree at Sun Yat-sen University, Guangzhou, China. His current research interests include autonomous driving and vehicle-road collaboration, deep learning and computer vision.
\end{IEEEbiography}

\vspace{35ex}
\begin{IEEEbiography}[{\includegraphics[width=1in,height=1.25in,clip,keepaspectratio]{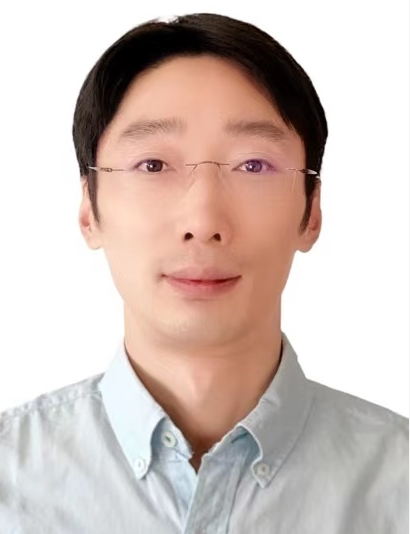}}]{Junzhou Chen}  received a PhD degree in Computer Science and Engineering from the Chinese University of Hong Kong in 2008. He is currently an associate professor at the School of Intelligent Systems Engineering in Sun Yat-sen University. His research interests focus on computer vision, machine learning, intelligent transportation and medical image processing.
\end{IEEEbiography}

\vspace{15ex}
\begin{IEEEbiography}[{\includegraphics[width=1in,height=1.25in,clip,keepaspectratio]{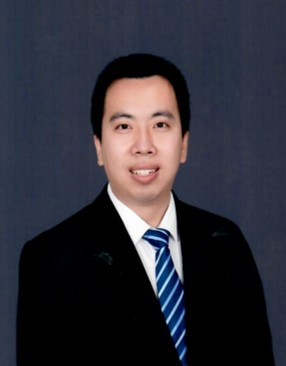}}]{Yilong Ren} received B.S. and Ph.D. degrees from Beihang University in 2010 and 2017, respectively. He is currently an Associate Professor at the School of Transportation Science and Engineering, Beihang University. His research interests include urban traffic operations and traffic control and simulation. 
\end{IEEEbiography}

\vspace{1ex}
\begin{IEEEbiography}[{\includegraphics[width=1in,height=1.25in,clip,keepaspectratio]{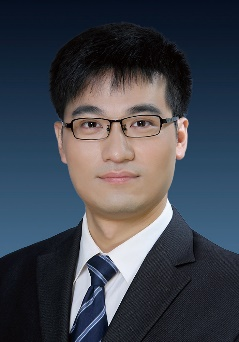}}]{Bolin Gao} received the B.S. and M.S. degrees in Vehicle Engineering from Jilin University, Changchun, China, in 2007 and 2009, respectively, and the Ph.D. degree in Vehicle Engineering from Tongji University, Shanghai, China, in 2013. He is now an associate research professor at the School of Vehicle and Mobility, Tsinghua University. His research interests include the theoretical research and engineering application of the dynamic design and control of Intelligent and Connected Vehicles, collaborative perception and tracking method in cloud control system, intelligent predictive cruise control system on commercial trucks with cloud control mode.
\end{IEEEbiography}

\vspace{25ex}
\begin{IEEEbiography}[{\includegraphics[width=1in,height=1.25in,clip,keepaspectratio]{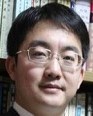}}] {Zhihan Lv} was born in Shandong, China, in 1984. He is currently a Professor at Qingdao University. Prior that he was a Research Associate at University College London (UCL). He has been also an assistant professor in Shenzhen Institutes of Advanced Technology, Chinese Academy of Sciences since 2012. He received his PhD from Paris7 University and Ocean University of China in 2012. He worked in CNRS (France) as Research Engineer, Umea University KTH Royal Institute of Technology (Sweden) as Postdoc Research Fellow, Fundacion FIVAN (Spain) as Experienced Researcher. He was a Marie Curie Fellow in European Union's Seventh Framework Programme LANPERCEPT. His research mainly focuses on Multimedia, Augmented Reality, Virtual Reality, Computer Vision, 3D Visualization \& Graphics, Serious Game, HCI, Bigdata, GIS. His research application fields widely range from everyday life to traditional research fields (i.e. geography, biology, medicine). He has completed several projects successfully on PC, Website, Smartphone and Smartglasses.
\end{IEEEbiography}

\end{document}